%% file: neurips_2022.tex
\documentclass{article}

\PassOptionsToPackage{numbers, compress}{natbib}
\usepackage[final]{neurips_2022}

\usepackage[utf8]{inputenc} % allow utf-8 input
\usepackage[T1]{fontenc}    % use 8-bit T1 fonts
\usepackage{hyperref}       % hyperlinks
\usepackage{url}            % simple URL typesetting
\usepackage{booktabs}       % professional-quality tables
\usepackage{amsfonts}       % blackboard math symbols
\usepackage{nicefrac}       % compact symbols for 1/2, etc.
\usepackage{microtype}      % microtypography
\usepackage{xcolor}         % colors

\usepackage{leftidx}
\usepackage{xcolor}

\usepackage{subfigure}
\usepackage{graphicx}
\usepackage{algorithm}
\usepackage[noend]{algpseudocode}
\usepackage{mdwtab}
\usepackage{eqparbox}
\usepackage{amsmath,amsfonts,amssymb}
\usepackage{hyperref}
\hypersetup{colorlinks=true,allcolors=blue}
\usepackage{caption} 
\usepackage{arevmath} % For math symbols
\usepackage{mathtools}% http://ctan.org/pkg/mathtools
\usepackage{array}
\usepackage{siunitx}
\usepackage{textgreek}
\usepackage{mathptmx}
\usepackage{comment}
\usepackage{booktabs, multirow} % for borders and merged ranges
\usepackage{soul}% for underlines
\usepackage{changepage,threeparttable} % for wide tables

\usepackage{tikz}
\usetikzlibrary{shapes.geometric, shapes.multipart, arrows, positioning, calc}
\usepackage{pgfplots}
\pgfplotsset{width=7.3cm,compat=newest}
\newcommand{\ours}{\textsc{RegCLR}}
\newcommand{\dataset}{EHRBank}

\title{\ours: A Self-Supervised Framework for \\ Tabular Representation Learning in the Wild}

\begin{document}

\input{sections/00_authors}

\maketitle

\input{sections/01_abstract}
\input{sections/10_introduction}

\input{sections/30_method}
\input{sections/40_exp_results}

\input{sections/20_related_work}
\input{sections/60_conclusion}

\begin{ack}
\input{sections/70_ack}
\end{ack}

% \section*{References}
\bibliography{neurips_2022}

%%%%%%%%%%%%%%%%%%%%%%%%%%%%%%%%%%%%%%%%%%%%%%%%%%%%%%%%%%%%

\appendix
\input{sections/100_appendix}

% \section{Appendix}

% Optionally include extra information (complete proofs, additional experiments and plots) in the appendix.
% This section will often be part of the supplemental material.

\end{document}

%% file: sections/00_authors.tex
% \author{Antiquus S.~Hippocampus, Natalia Cerebro \& Amelie P. Amygdale \thanks{ Use footnote for providing further information
% about author (webpage, alternative address)---\emph{not} for acknowledging
% funding agencies.  Funding acknowledgements go at the end of the paper.} \\
% Department of Computer Science\\
% Cranberry-Lemon University\\
% Pittsburgh, PA 15213, USA \\
% \texttt{\{hippo,brain,jen\}@cs.cranberry-lemon.edu} \\
% \And
% Ji Q. Ren \& Yevgeny LeNet \\
% Department of Computational Neuroscience \\
% University of the Witwatersrand \\
% Joburg, South Africa \\
% \texttt{\{robot,net\}@wits.ac.za} \\
% \AND
% Coauthor \\
% Affiliation \\
% Address \\
% \texttt{email}
% }

\author{%
  Weiyao Wang\thanks{equal technical contribution} \\
  Johns Hopkins University \\
  \texttt{wwang121@jhu.edu} \\
  \And
  Byung-Hak Kim\(^{*,}\)\thanks{project lead}\\
  AKASA, Inc. \\
  \texttt{hak.kim@AKASA.com} \\
  \And
  Varun Ganapathi \\
  AKASA, Inc. \\
  \texttt{varun@AKASA.com} \\
}

%% file: sections/01_abstract.tex
\begin{abstract}
Recent advances in self-supervised learning (SSL) using large models to learn visual representations from natural images are rapidly closing the gap between the results produced by fully supervised learning and those produced by SSL on downstream vision tasks. Inspired by this advancement and primarily motivated by the emergence of tabular and structured document image applications, we investigate which self-supervised pretraining objectives, architectures, and fine-tuning strategies are most effective. To address these questions, we introduce \ours,~a new self-supervised framework that combines contrastive and regularized methods and is compatible with the standard Vision Transformer~\citep{Dosovitskiy21} architecture. Then, \ours~is instantiated by integrating masked autoencoders~\citep{He22} as a representative example of a contrastive method and enhanced Barlow Twins as a representative example of a regularized method with configurable input image augmentations in both branches. Several real-world table recognition scenarios (e.g., extracting tables from document images), ranging from standard Word and Latex documents to even more challenging electronic health records (EHR) computer screen images, have been shown to benefit greatly from the representations learned from this new framework, with detection average-precision (AP) improving relatively by 4.8\% for Table, 11.8\% for Column, and 11.1\% for GUI objects over a previous fully supervised baseline on real-world EHR screen images. 
\end{abstract}
% \vspace{-2ex} 

%% file: sections/10_introduction.tex
\section{Introduction}
\label{sec:introduction}

Many self-supervised learning (SSL) methods for learning visual representation without supervision (or labels) have been proposed in recent years~\citep{zhang2016colorful, noroozi2016unsupervised, gidaris2018unsupervised, Chen20, grill2020bootstrap, zbontar2021barlow, He22}. Indeed, the potentially practical benefits of avoiding the high cost of human annotations and moving away from human-defined and language-dependent label categories strongly motivate this direction. In particular, the most common framework is to use abundant \textit{unlabeled} images to pretrain rich visual representations and then transfer them to supervised downstream tasks with much fewer annotated ones. 

Tables are the most prevalent means of representing and communicating structured data in a wide range of document images, such as financial statements, scientific papers, and electronic medical health documents. Despite the explosive growth of the number of these document images~\citep{Vashisth22}, most SSL approaches have so far been proposed in the context of the natural image domain, with little attention paid to tabular and structured document image domains. Even the most advanced systems (e.g.,~\citet{Prasad20,Agarwal21}) are largely dependent on object detection models that have been trained on human-labeled natural images. 

Table objects are a compact representation that humans can easily understand. However, this is not true for machines since, unlike classic object detection classes, they might have widely disparate sizes, types, styles, and aspect ratios. In other words, table structure varies greatly between document domains (e.g., Word vs GUI screen), and a large variety of table styles are feasible even within the same document format (e.g., borderless vs bordered). 

More importantly, object detection models trained for natural images (e.g., Faster R-CNN or YOLO) retrained for table can work well locally while ignoring the overall style of tables that is important to the table~\citep{Burdick20}. In that regard, very few works have started exploring SSL approaches to the problem of tabular rich document image domains. DiT~\citep{li2022dit}, for example, directly employs BERT-style BEiT~\citep{bao2021beit} to pretrain in a self-supervised manner on IIT-CDIP dataset~\citep{lewis2006building} of 42 million document images and then fine-tune on a couple of classification and detection tasks. 

However, to the best of our knowledge, there are several open research problems in developing SSL methods, particularly for tabular and structured document image domains, that require further study. Finding simple but effective self-supervised pretraining objectives, architectures, and fine-tuning strategies are examples of this. Keeping those questions in mind, our paper makes two main contributions:
 \begin{itemize}
    \item First, we introduce \ours, a new self-supervised framework that combines contrastive-based masked image modeling (MIM)\footnote{We follow Yan LeCun's broader classification of contrastive methods introduced in~\citep{LeCun22}, which includes recent standard approaches (e.g., SimCLR~\citep{Chen20}), MIM family (e.g., MAE~\citep{He22}), and even Generative Adversarial Networks.}  and regularized methods and is Vision Transformer (ViT) compatible. It is then instantiated by coupling masked autoencoders (MAE) and enhanced Barlow Twins (eBT) with configurable image augmentations; in principle, the MAE loss is augmented with the eBT loss as a regularization (prior) term. In fact, this simple SSL training loss objective is highly effective in reducing the capacity of learned/trained latent representation, especially in the balanced direction of MAE and eBT loss. MAE loss chooses to ignore irrelevant details, while eBT loss pushes relevant details to latent vectors (i.e., minimizing the ignored details), which is highly important for tabular representation learning.
    
    \item Second, we pretrain \ours~on two distinct domains of document image datasets with rich tabular structures, the publicly accessible TableBank~\citep{li2019tablebank} and more challenging internal EHRBank, and then validate the higher sample efficiency of the learned tabular representations in finetuning with TableBank and the superior detection performance with EHRBank in various practical downstream settings. A recent MIMDet~\citep{fang2022unleashing} framework is used to extract multiscale features from the pretrained ViT encoder and decoder to use as a detection backbone without the need for architecture redesign or changes. By leveraging MIMDet, \ours~outperforms the latest and purely supervised convolutional neural network (CNN)  baseline by large margins. 
\end{itemize}

%% file: sections/30_method.tex
\section{\ours~Description}
\label{sec:method}
\subsection{Information Theoretic Formulation}
\textbf{Problem Motivation and Notations} In order to build a new unified SSL framework that incorporates the best of both worlds of MIM-based contrastive and regularized self-supervised approaches, we adopt an information-theoretic perspective to pose the problem. The main question is well expressed in the following quote from Yan Lecun's recent positional paper~\citep[p.~23]{LeCun22}.  

\begin{quote}[\cite{LeCun22}]{}
---It is important to note that contrastive and regularized methods are not incompatible with each other, and can be used \textit{simultaneously} on the same model. How would regularized methods apply to the standard architectures?---
\end{quote}

To begin with, let \(X\) be the input image, \((X_1,X_2,X_3)\) be the augmented or varied views of the image, and \((Y_1,Y_2,Y_3)\) be the corresponding hidden/latent representation. Our goal is to learn the best representation \(Y\) that retains as much information about the input image \(X\) as possible given a reasonable representation constraint. The information bottleneck (IB) framework~\citep{Slonim06} can be a good constraint to use in this case. To use the IB principle, denote \(n\) be the number of samples, \(d\) be the feature dimension, and \(Z_2, Z_3\in\mathbb{R}^{n \times d}\) be the projected features. 

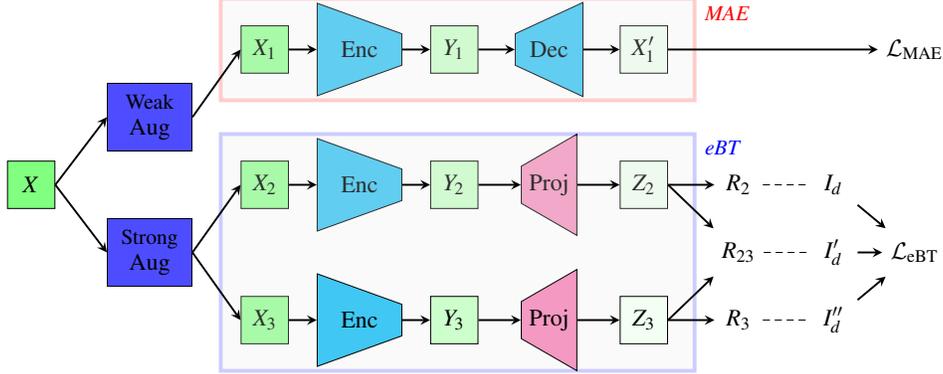
\begin{figure*}[ht]
    \centering
    \scalebox{.9}{\input{figures/arch.tikz}}
    \caption{\textbf{\ours's self-supervised pretraining}. \ours~is consisted of two branches: MAE and eBT. MAE branch uses weak augmentation and masking to obtain \(X_1\) and then follows procedure introduced in MAE~\citep{He22} to compute reconstructed loss for unmasked patches as \(\mathcal{L_{\mathrm{MAE}}}\) to obtain \(X_1'\). In the eBT branch, strong augmentation is applied twice to get \(X_2\) and \(X_3\). Through the same ViT encoder used in the MAE branch, the encoded images are then projected into features \(Z_2\) and \(Z_3\). The proposed loss \(\mathcal{L_{\mathrm{eBT}}}\) computes three correlation matrices between \(Z_2\) and \(Z_3\), attempting to make each matrix near to an identity matrix. \ours's overall self-supervised pretraining is performed by jointly minimizing \(\mathcal{L_{\mathrm{MAE}}}\) and \(\mathcal{L_{\mathrm{eBT}}}\) through optimizing both ViT encoder and decoder.}
    \label{fig:Arch}
    \vspace{-1ex} 
\end{figure*}

The role of the IB constraint is to compress \(X_2\) into \(Z_2\) while preserving the information about \(Z_3\), and similarly \(X_3\) into \(Z_3\) while keeping the information about \(Z_2\). To put it another way, the constraint seeks to compress \(X_2\) and \(X_3\) as much as possible, while also making \(Z_2\) and \(Z_3\) as informative about each other as possible. For simplicity, we assume that \(Z_2\) and \(Z_3\) are two jointly multivariate Gaussian variables with zero means and covariance matrices of \(K_2,K_3\in\mathbb{R}^{d \times d}\) that are full rank. Note that \(|K|\) denotes the determinant of \(K\), \(\lambda_{1}, \lambda_{2}, ... , \lambda_{n}\) are the eigenvalues of \(K\), \(||\cdot||_{F}\) is the Frobenius norm, and \(h\) is the differential entropy.

\textbf{Training Loss Design} From an information theoretic perspective, the problem can be written as an optimization problem in particular as:
\begin{equation}
\begin{split}
  \arg\max_{\theta} I(X_1;Y_1|\theta)-I(Z_2;X_2|\theta)-I(Z_3;X_3|\theta)+\alpha I(Z_2;Z_3|\theta). 
\end{split}
\end{equation}
Equivalently, if we omit \(\theta\) for brevity, the target function \(\mathcal{L}\) to minimize can be expressed as:
\begin{equation}\label{eqn:full_loss}
\begin{split}
    \mathcal{L} &= \underbrace{h(X_1|Y_1)}_\textrm{\(\mathcal{L}_1\)}+\underbrace{(1-\alpha)h(Z_2)+(1-\alpha)h(Z_3)+\alpha h(Z_2,Z_3)}_\textrm{\(\mathcal{L}_2\)}, 
\end{split}    
\end{equation}
this equality follows from the fact that \(h(X_1)\) is a constant when the input distribution is fixed and \(h(Z_2|X_2)=h(Z_3|X_3)=0\). It has previously been shown \citep{Vincent08} that minimizing the expected reconstruction error equates to maximizing a lower bound on the first loss term  \(\mathcal{L}_1\) in (\ref{eqn:full_loss}). This holds true even when \(Y\) is a function of a corrupted input (e.g., masked out image). Therefore, for example, the mean squared error (MSE) between the augmented image \(X_1\) and the reconstructed version \(X_1'\) denoted as \(\mathcal{L_{\mathrm{MAE}}}\) can be used as the equivalent loss function of \(\mathcal{L}_1\).

We can further express the remaining three loss terms \(\mathcal{L}_2\) in (\ref{eqn:full_loss}) as\footnote{Recall that the entropy of an \(d\) dimensional Gaussian variable is \(h(Z)=\frac{1}{2}\log\Big(\left(2\pi e\right)^d|K|\Big)\).}
\begin{equation}\label{eqn:loss2}
\begin{split}
    \log\big(|K_2|\big)+\log\big(|K_3|\big)+\beta\log\big(|K_{23}|\big), 
\end{split}    
\end{equation}
where \(\beta\triangleq\big(\frac{\alpha}{1-\alpha}\big)\). Then, \(\log\big(|K|\big)\) can be upper bounded to be computationally convenient to use as follows: 
\begin{equation}\label{eqn:kbound}
\begin{split}
    \log\big(|K|\big) &= \log\Bigg(\prod_{i=1}^{n}\lambda_{i}\Bigg) = \sum_{i=1}^{n}\log\lambda_{i} < \sum_{i=1}^{n} \lambda_{i}^{2} = ||K||_{F}^{2}.
\end{split}    
\end{equation}
Combining (\ref{eqn:kbound}) with (\ref{eqn:loss2}), we obtain the further simplified version of \(\mathcal{L}_2\) as
\begin{equation}
\begin{split}\label{eqn:simpler_loss2}
    ||K_2||_{F}^{2}+||K_3||_{F}^{2}+\beta||K_{23}||_{F}^{2}.
\end{split}    
\end{equation}

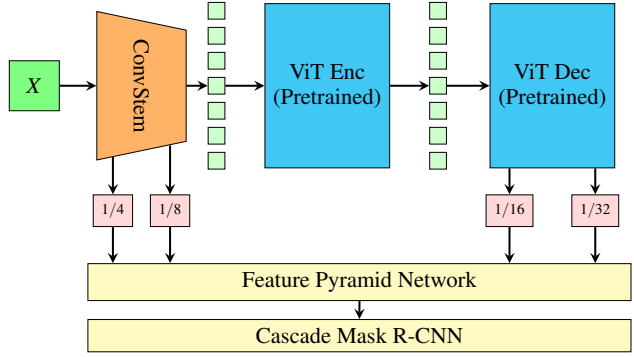
\begin{figure*}[ht]
  \centering 
  \scalebox{.95}{\input{figures/mimdet.tikz}}
  \caption{\textbf{\ours's~detection via ViT backbone}. The MIMDet architecture is leveraged by~\ours, as described in Section~\ref{ssec:framework}, to transfer both the pretrained ViT encoder and decoder to detection with Cascade Mask R-CNN without requiring the ViT network redesign. In MIMDet, a randomly initialized ConvStem replaces the pretrained large kernel PatchStem, and the ConvStem's intermediate features can directly be used as higher resolution inputs for a standard feature pyramid network.}
  \label{fig:mimdet}
\end{figure*}

\begin{figure}[!ht]
  \centering 
  \includegraphics[width=5.5in]{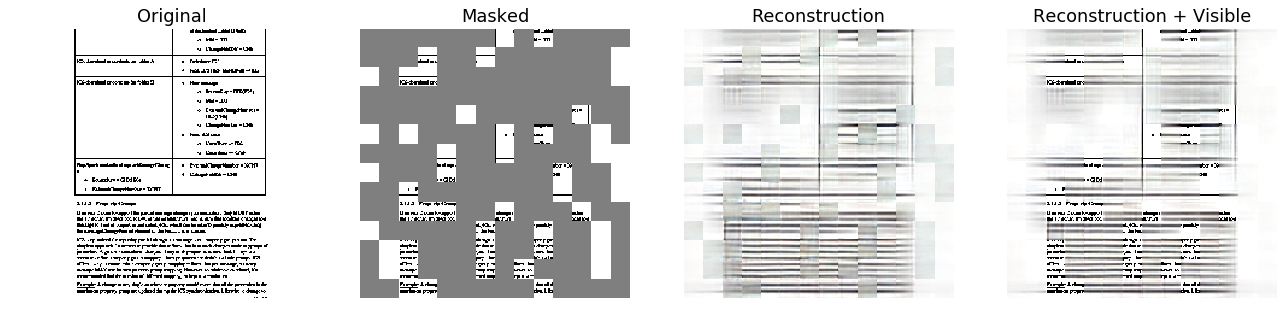}
  \caption{\textbf{Visualization of the MAE branch's reconstruction} for a TableBank Word test set example (with a masking ratio of 75\%). The MAE arm reconstructs the table's structure as well as the most of the masked boundaries, demonstrating that it can learn tabular semantic information from only unmasked patches while ignoring irrelevant document details (the eBT arm, on the other hand, is anticipated to keep the ignored details to a minimum).}
  \label{fig:mae-vis}
  \vspace{-1ex} 
\end{figure}

Since \(Z_2\) and \(Z_3\) are zero means, the cross-covariance matrix \(K_{23}\) becomes the cross-correlation matrix \(R_{23}\) assuming normalization without loss of generality. Similarly, auto-covariance matrices \(K_{Z_2}\) and \(K_3\) become the auto-correlation matrices \(R_2\) and \(R_3\), respectively. Because there is a trivial solution of zero matrices when minimizing (\ref{eqn:simpler_loss2}), we can circumvent this trivial solution by urging the diagonal terms to be \(1\) and all the off-diagonal terms to be zero. Hence, the resulting objective denoted as \(\mathcal{L_{\mathrm{eBT}}}\) is given by 
\begin{equation}
\begin{split}
    \mathcal{L_{\mathrm{eBT}}} &= \nu_{1}\sum_{i}\Big(1-R_{2,ii}\Big)^{2}+\nu_{1}\sum_{i}\Big(1-R_{3,ii}\Big)^{2}+\nu_{2}\sum_{i}\Big(1-R_{23,ii}\Big)^{2}\\
    &+\mu_{1}\sum_{i}\sum_{j\neq i}\Big(R_{2,ij}\Big)^{2}+\mu_{1}\sum_{i}\sum_{j\neq i}\Big(R_{3,ij}\Big)^{2}+\mu_{2}\sum_{i}\sum_{j\neq i}\Big(R_{23,ij}\Big)^{2},
\end{split}    
\end{equation}
where \(\nu_1\), \(\nu_2\), \(\mu_1\) and \(\mu_2\) are hyper-parameters controlling diagonal and off-diagonal terms of each matrix. We use \(\mu_1\) = 0.5, \(\mu_2\) = 1.0, \(\nu_1\) = 0.00255 and \(\nu_2\) = 0.0051 to balance between auto-correlation and cross-correlation terms during training. After introducing \(\mathcal{L_{\mathrm{MAE}}}\) and \(\mathcal{L_{\mathrm{eBT}}}\), the overall training loss is now straightforward to compose as: 
\begin{equation}\label{eqn:total_loss}
\begin{split}
    \mathcal{L_{\mathrm{\ours}}} &= \mathcal{L_{\mathrm{MAE}}} + \mathcal{L_{\mathrm{eBT}}}. 
\end{split}    
\end{equation}

    \begin{table}[!ht]
        \centering
        \caption{\textbf{TableBank and \dataset~dataset statistics}. TableBank includes Word and Latex document image sets, while internally curated \dataset~has EHR screen images in the form of tables, table columns, and GUI elements sets. Because labeling all GUI elements in the EHR images is more costly, we have fewer samples in the GUI dataset. Note that the \textit{unlabeled} \dataset~Screenshot dataset (shown in the last row) is used only for pretraining.}\label{tab:dataset_size}
        \footnotesize
        \begin{tabular}{lrrrr}\toprule
        &Train &Val &Test \\ \midrule
        TableBank Word &73,383 &2,735 &2,281 \\
        TableBank Latex &187,199 &7,265 &5,719 \\ \midrule
        \dataset~Table &1,917 &411 &209 \\
        \dataset~Column &1,194 &255 &208 \\
        \dataset~GUI &157 &38 &157 \\ \midrule
        \dataset~Screenshot (\textit{unlabeled}) &28,121 & & \\
        \bottomrule
        \end{tabular}
        \label{tab:datasets}
        \vspace{-1ex} 
    \end{table}

\subsection{\ours~Framework}
\label{ssec:framework}
\textbf{Self-Supervised Pretraining} 
One possible instantiation is to use the new objective, \(\mathcal{L_{\mathrm{\ours}}}\) in (\ref{eqn:total_loss}), for self-supervised pretraining, and build the model with two branches, MAE and eBT, as shown in Figure~\ref{fig:Arch}. For the MAE branch, we choose to apply weak augmentation to the input image and then randomly mask out the selected patches. Only unmasked patches are fed into the ViT encoder, and subsequently the masked patches are reconstructed by the Vit decoder using the MSE calculated over the masked patches as the loss function \(\mathcal{L_{\mathrm{MAE}}}\) per the original MAE design (see Figure~\ref{fig:mae-vis} for an example of image visualization). 

Secondly, the eBT branch operates on the cross embedding of the input image's two strongly augmented versions \(X_2\) and \(X_3\). In contrast to the previous instantiation of the IB principle presented in BT loss~\citet{zbontar2021barlow}, which computes only the cross-correlation matrix, \(\mathcal{L_{\mathrm{eBT}}}\) computes a cross-correlation matrix as well as two auto-correlation matrices bringing these three matrices close to the identity ones. We believe that this is a more natural and fundamental improvement over BT loss because augmented images typically carry a significant amount of input image information, implying that the conditional independence assumption of the BT loss ~\citep[Figure 6]{zbontar2021barlow} that leads to minimizing \(I(Z_2;X_2|\theta)\) while maximizing \(I(Z_3;X_3|\theta)\) is suboptimal.  

\textbf{Detection via ViT Backbone} 
For detection, as illustrated in Figure~\ref{fig:mimdet}, we combine a ViT encoder and decoder pretrained in a self-supervised manner with MIMDet~\citep{fang2022unleashing} to serve as the detection backbone and leverage Cascade Mask R-CNN~\citep{Cai_2019}, which is the common architecture in supervised state-of-the-art systems (e.g.,~\citet{Prasad20,Agarwal21}). Compared to previous representative approaches of adapting vanilla ViT for object detection (e.g.~\citet{li2021benchmarking}), MIMDet replaces the pretrained patchify stem (PatchStem) with a compact convolutional stem (ConvStem) without further upsampling or redesigns, resulting in a ConvNet-ViT hybrid multi scale feature extractor that requires far fewer epochs in the fine-tuning procedure.

%% file: figures/arch.tikz
\tikzset{
img/.style= {draw, fill=green!50, rectangle, minimum width=0.7cm, minimum height=0.7cm, node distance=1cm},
aug/.style= {draw, fill=blue!70, rectangle, minimum width=1.25cm, minimum height=1.0cm, node distance=1cm},
aug_out/.style= {draw, fill=green!40, rectangle, minimum width=0.7cm, minimum height=0.7cm, node distance=1cm},
enc_t/.style= {draw, fill=cyan!60, trapezium, rotate=270, trapezium stretches=true, minimum width=1.0cm, minimum height=1.25cm, node distance=1cm},
dec_t/.style= {draw, fill=cyan!60, trapezium, rotate=90, trapezium stretches=true, minimum width=1.0cm, minimum height=1.0cm, node distance=1cm},
rep/.style= {draw, fill=green!20, rectangle, minimum width=0.7cm, minimum height=0.7cm, node distance=1cm},
proj_t/.style= {draw, fill=magenta!50, trapezium, rotate=90, trapezium stretches=true, minimum width=1.0cm, minimum height=0.75cm, node distance=1cm},
feat/.style= {draw, fill=green!5, rectangle, minimum width=0.7cm, minimum height=0.7cm, node distance=1cm},
corr/.style= {fill=white, rectangle, minimum width=0.7cm, minimum height=0.7cm, node distance=1cm},
box_mae/.style = {draw=red, ultra thick, fill=black!10, opacity=.2, rectangle, minimum width=7.0cm, minimum height=1.5cm},
box_ebt/.style = {draw=blue, ultra thick, fill=black!10, opacity=.2, rectangle, minimum width=7.0cm, minimum height=3.5cm},
}

\tikzstyle{arrow} = [thick,->,>=stealth]

\begin{tikzpicture}[node distance=1.0cm]
    \tikzstyle{every node}=[font=\normalsize]
    \node[text=black] (in) [img] {\(X\)};
    \node[text=black, align=center] (weak_aug)[aug, right of=in, xshift=0.75cm, yshift=1.0cm] {\small Weak\\Aug};
    \node[text=black, align=center] (strong_aug)[aug, right of=in, xshift=0.75cm, yshift=-1.0cm] {\small Strong\\Aug};

    \node (x1) [aug_out, right of=weak_aug, xshift=0.7cm, yshift=1.0cm] {\(X_1\)};
    \node (x2) [aug_out, right of=strong_aug, xshift=0.7cm, yshift=1.0cm] {\(X_2\)};
    \node (x3) [aug_out, right of=strong_aug, xshift=0.7cm, yshift=-1.0cm] {\(X_3\)};

    \node (enc1) [enc_t, above of=x1, xshift=0.0cm, yshift=0.4cm] {\normalsize \rotatebox{90}{Enc}};
    \node (rep1) [rep, right of=enc1, xshift=0.4cm] {\(Y_{1}\)};
    \node (dec1) [dec_t, right of=rep1, xshift=-1.0cm, yshift=-1.4cm] {\normalsize \rotatebox{270}{Dec}};
    \node (rec) [feat, right of=dec1, xshift=0.4cm] {\(X_1'\)};

    \node (mae)[box_mae, above of=x1, xshift=2.85cm, yshift=-1.0cm]{};
    \node[below right, text=red] at (mae.north east) {\small\textit{MAE}};
    
    \node (enc2) [enc_t, above of=x2, xshift=0.0cm, yshift=0.4cm] {\normalsize \rotatebox{90}{Enc}};
    \node (rep2) [rep, right of=enc2, xshift=0.4cm] {\(Y_{2}\)};
    \node (proj2) [proj_t, right of=rep2, xshift=-1.0cm, yshift=-1.4cm] {\normalsize \rotatebox{270}{Proj}};
    \node (out2) [feat, right of=proj2, xshift=0.4cm] {\(Z_{2}\)};
    
    \node (ebt)[box_ebt, above of=x2, xshift=2.85cm, yshift=-2.0cm]{};
    \node[below right, text=blue] at (ebt.north east) {\small\textit{eBT}};

    \node (enc3) [enc_t, above of=x3, xshift=0.0cm, yshift=0.4cm] {\normalsize \rotatebox{90}{Enc}};
    \node (rep3) [rep, right of=enc3, xshift=0.4cm] {\(Y_{3}\)};
    \node (proj3) [proj_t, right of=rep3, xshift=-1.0cm, yshift=-1.4cm] {\normalsize\rotatebox{270}{Proj}};
    \node (out3) [feat, right of=proj3, xshift=0.4cm] {\(Z_{3}\)};

    \node (corr1) [corr, right of=out2, xshift=0.4cm] {\(R_{2}\)};
    \node (corr2) [corr, below of=corr1, xshift=0.0cm] {\(R_{23}\)};
    \node (corr3) [corr, right of=out3, xshift=0.4cm] {\(R_{3}\)};

    \node (iden1) [corr, right of=corr1, xshift=0.4cm] {\(I_{d}\)};
    \node (iden2) [corr, below of=iden1, xshift=0.0cm] {\(I_{d}'\)};
    \node (iden3) [corr, below of=iden2, xshift=0.0cm] {\(I_{d}''\)};
    
    \draw [arrow] (in.east) -- (weak_aug.west);
    \draw [arrow] (in.east) -- (strong_aug.west);
    \draw [arrow] (weak_aug.east) -- (x1.west);
    \draw [arrow] (strong_aug.east) -- (x2.west);
    \draw [arrow] (strong_aug.east) -- (x3.west);

    \draw [arrow] (x1.east) -- (enc1.south);
    \draw [arrow] (enc1) -- (rep1);
    \draw [arrow] (rep1) -- (dec1);
    \draw [arrow] (dec1) -- (rec);
    
    \draw [arrow] (x2) -- (enc2);
    \draw [arrow] (enc2) -- (rep2);
    \draw [arrow] (rep2) -- (proj2);
    \draw [arrow] (proj2) -- (out2);
    
    \draw [arrow] (x3) -- (enc3);
    \draw [arrow] (enc3) -- (rep3);
    \draw [arrow] (rep3) -- (proj3);
    \draw [arrow] (proj3) -- (out3);
    
    \draw [arrow] (out2) -- (corr1);
    \draw [arrow] (out2.east) -- (corr2);
    \draw [arrow] (out3.east) -- (corr2);
    \draw [arrow] (out3) -- (corr3);
    
    \draw [densely dashed] (corr1) -- (iden1); 
    \draw [densely dashed] (corr2) -- (iden2);
    \draw [densely dashed] (corr3) -- (iden3);
    
    \node (loss1) [corr, right of=rec, xshift=3.0cm] {\(\mathcal{L_{\mathrm{MAE}}}\)};
    \node (loss2) [corr, right of=iden2, xshift=0.2cm] {\(\mathcal{L_{\mathrm{eBT}}}\)};

    \draw [arrow] (rec) -- (loss1);
    \draw [arrow] (iden1) -- (loss2);
    \draw [arrow] (iden2) -- (loss2);
    \draw [arrow] (iden3) -- (loss2);
    
\end{tikzpicture}

%% file: figures/mimdet.tikz
\tikzset{
img/.style= {draw, fill=green!50, rectangle, minimum width=0.7cm, minimum height=0.7cm, node distance=1cm},
conv_t/.style= {draw, fill=orange!60, trapezium, rotate=-90, trapezium stretches body, minimum width=1.0cm, minimum height=1.25cm, node distance=1cm},
enc/.style= {draw, fill=cyan!60, rectangle, minimum width=0.7cm, minimum height=2.3cm, node distance=0.05cm},
dec/.style= {draw, fill=cyan!60, rectangle, minimum width=0.7cm, minimum height=2.3cm, node distance=1cm},
convert/.style= {draw, fill=pink!60, rectangle, minimum width=0.1cm, minimum height=0.1cm, node distance=.75cm},
patch/.style= {draw, fill=green!20, rectangle, minimum width=0.1cm, minimum height=0.1cm, node distance=.35cm},
fpn/.style= {draw, fill=yellow!30, rectangle, minimum width=7.6cm, minimum height=0.1cm, node distance=.75cm},
rcnn/.style= {draw, fill=yellow!30, rectangle, minimum width=7.6cm, minimum height=0.1cm, node distance=.75cm},
}

\tikzstyle{arrow} = [thick,->,>=stealth]

\begin{tikzpicture}[node distance=1.0cm]
    \tikzstyle{every node}=[font=\small]
    \node[text=black] (in) [img] {\(X\)};
    \node(conv)[conv_t, above of=in, xshift=0.0cm, yshift=0.5cm] {ConvStem};
    
    \node (p1) [patch, right of=conv, xshift=0.7cm, yshift=0.0cm] {};
    \node (p2) [patch, above of=p1, xshift=0.0cm, yshift=0.0cm] {};
    \node (p3) [patch, above of=p2, xshift=0.0cm, yshift=0.0cm] {};
    \node (p4) [patch, above of=p3, xshift=0.0cm, yshift=0.0cm] {};
    \node (p5) [patch, below of=p1, xshift=0.0cm, yshift=0.0cm] {};
    \node (p6) [patch, below of=p5, xshift=0.0cm, yshift=0.0cm] {};
    \node (p7) [patch, below of=p6, xshift=0.0cm, yshift=0.0cm] {};
    
    \node [align=center](enc)[enc, right of=p1, xshift=1.5cm, yshift=0.0cm] {ViT Enc\\(Pretrained)};

    \node (p8) [patch, right of=enc, xshift=1.2cm, yshift=0.0cm] {};
    \node (p9) [patch, above of=p8, xshift=0.0cm, yshift=0.0cm] {};
    \node (p10) [patch, above of=p9, xshift=0.0cm, yshift=0.0cm] {};
    \node (p11) [patch, above of=p10, xshift=0.0cm, yshift=0.0cm] {};
    \node (p12) [patch, below of=p8, xshift=0.0cm, yshift=0.0cm] {};
    \node (p13) [patch, below of=p12, xshift=0.0cm, yshift=0.0cm] {};
    \node (p14) [patch, below of=p13, xshift=0.0cm, yshift=0.0cm] {};

    \node [align=center](dec)[dec, right of=p8, xshift=0.6cm, yshift=0.0cm] {ViT Dec\\(Pretrained)};

    \node (s1) [convert, below of=conv, xshift=-0.4cm, yshift=-1.0cm] {\tiny \(1/4\)};
    \node (s2) [convert, below of=conv, xshift=0.4cm, yshift=-1.0cm] {\tiny \(1/8\)};
    \node (s3) [convert, below of=dec, xshift=-0.6cm, yshift=-1.0cm] {\tiny \(1/16\)};
    \node (s4) [convert, below of=dec, xshift=0.6cm, yshift=-1.0cm] {\tiny \(1/32\)};
    
    \node (fpn) [fpn, below of=enc, xshift=0.45cm, yshift=-2.0cm] {Feature Pyramid Network};
    \node (rcnn) [rcnn, below of=fpn, xshift=0.0cm, yshift=-0.0cm] {Cascade Mask R-CNN};

    \draw [arrow] (in.east) -- (conv);
    \draw [arrow] (conv) -- (p1);
    \draw [arrow] (p1) -- (enc);
    \draw [arrow] (enc) -- (p8);
    \draw [arrow] (p8) -- (dec);
    
    \draw [arrow] ([xshift=-0.4cm,yshift=-0.08cm]conv.east) -- (s1.north);
    \draw [arrow] ([xshift=0.4cm,yshift=0.08cm]conv.east) -- (s2.north);
    \draw [arrow] ([xshift=-0.6cm]dec.south) -- (s3.north);
    \draw [arrow] ([xshift=0.6cm]dec.south) -- (s4.north);
    
    \draw [arrow] (s1.south) -- ([xshift=-3.45cm]fpn.north);
    \draw [arrow] (s2.south) -- ([xshift=-2.65cm]fpn.north);
    \draw [arrow] (s3.south) -- ([xshift=2.1cm]fpn.north);
    \draw [arrow] (s4.south) -- ([xshift=3.3cm]fpn.north);
    
    \draw [arrow] (fpn) -- (rcnn);
 
\end{tikzpicture}

%% file: sections/40_exp_results.tex
\section{Experiments}
\label{sec:expresults}
\subsection{Datasets}
\begin{figure*}[!ht]
      \centering 
      \subfigure{
        \includegraphics[width=1.25in]{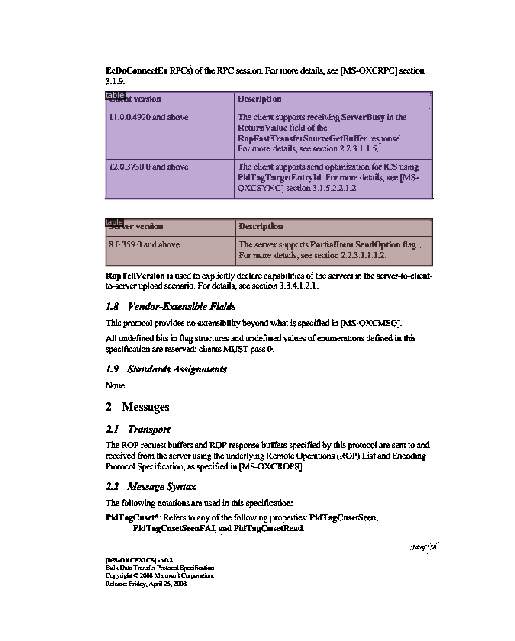}
      }
      \subfigure{
        \includegraphics[width=1.25in]{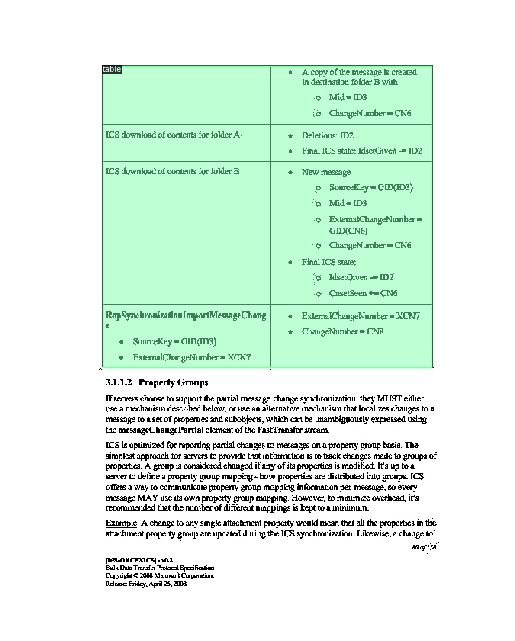}
      }
      \subfigure{
        \includegraphics[width=1.2in]{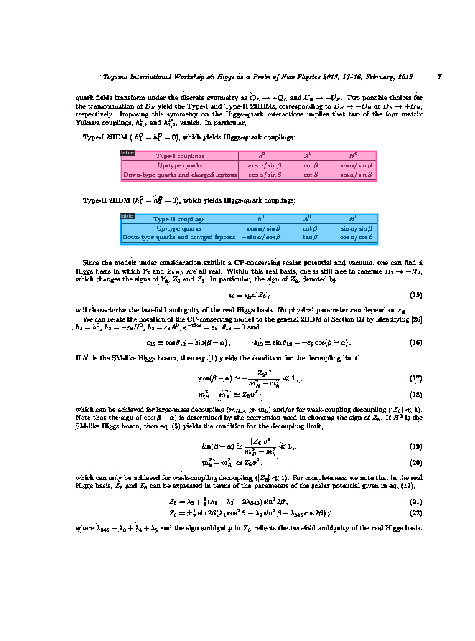}
      }
      \subfigure{
        \includegraphics[width=1.25in]{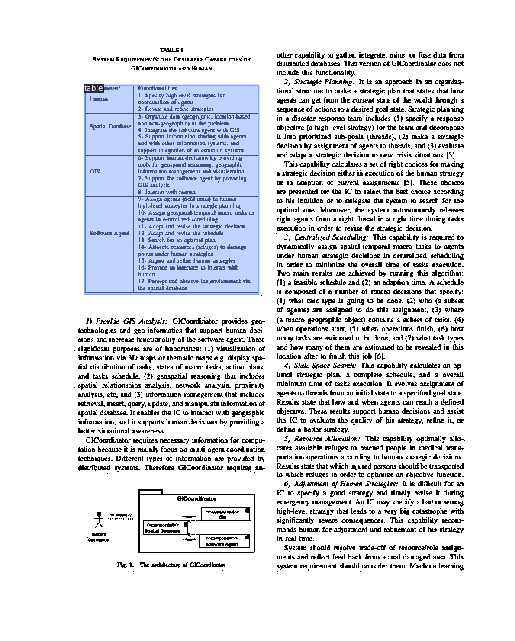}
      }
      \vspace{-2ex} 
      \setcounter{subfigure}{0}
      \subfigure[]{
        \includegraphics[width=1.25in]{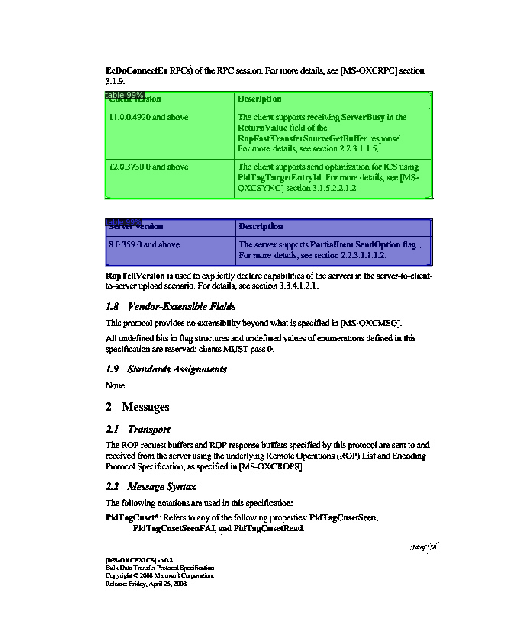}
      }
      \subfigure[]{
        \includegraphics[width=1.25in]{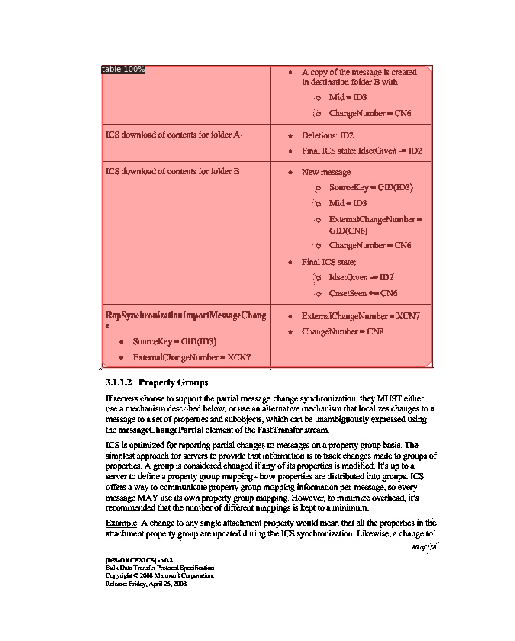}
      }
      \subfigure[]{
        \includegraphics[width=1.2in]{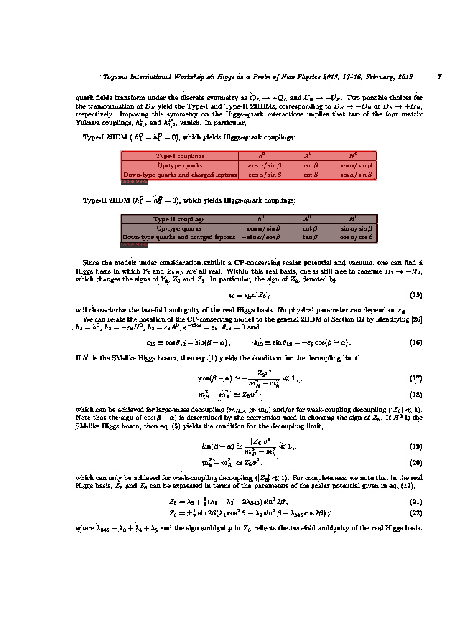}
      }
      \subfigure[]{
        \includegraphics[width=1.25in]{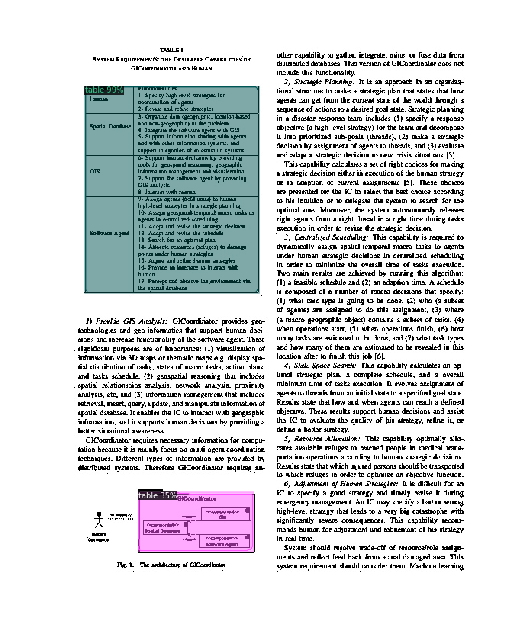}
      }
      \caption{\textbf{TableBank test set examples} (a-b) Word test set and (c-d) Latex test set. The top row depicts examples with annotated boundaries, while the bottom row presents the corresponding detection predictions by~\ours~.} 
      \label{fig:tb-example}
      \vspace{-1ex} 
    \end{figure*}

\textbf{TableBank Dataset}
TableBank is a publicly available image-based table dataset for Word and Latex documents on the Internet. It contains 417K high-quality tables labeled with weak supervision split across Word and Latex sets. Note that the Word document set contains English, Chinese, Japanese, Arabic, and other languages, whereas the Latex document set is primarily in English. Details of the data split can be found in Table~\ref{tab:datasets} and one can refer to the top row of Figure~\ref{fig:tb-example} for image samples with annotated bounding boxes. 
    
\begin{figure}[!ht]
      \centering 
      \subfigure[]{
        \includegraphics[width=2.25in]{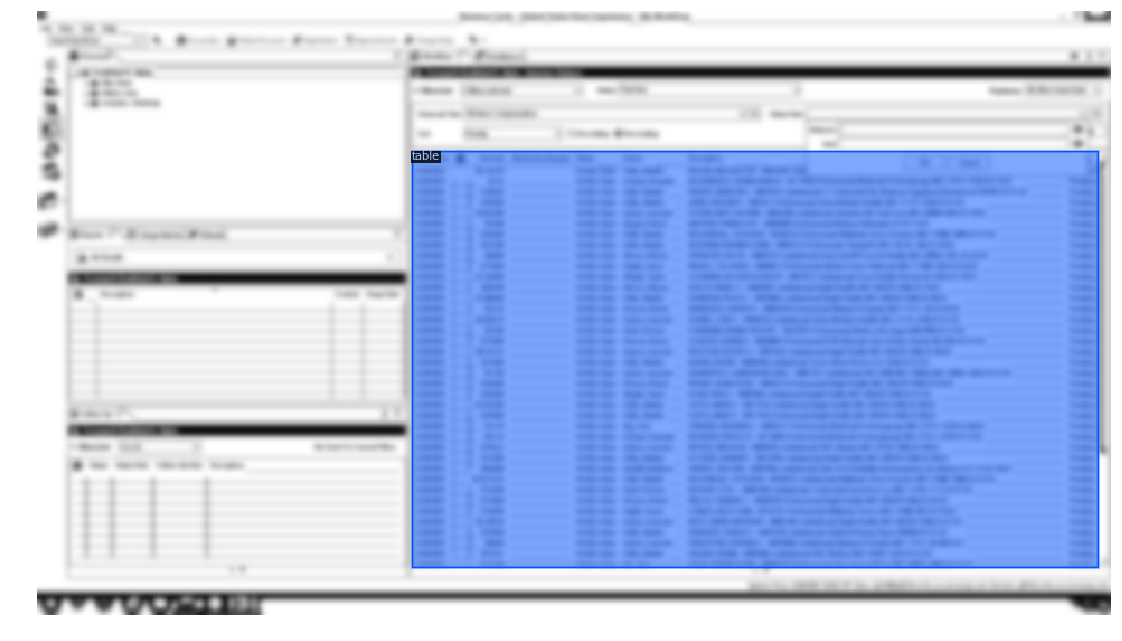}
        \includegraphics[width=2.25in]{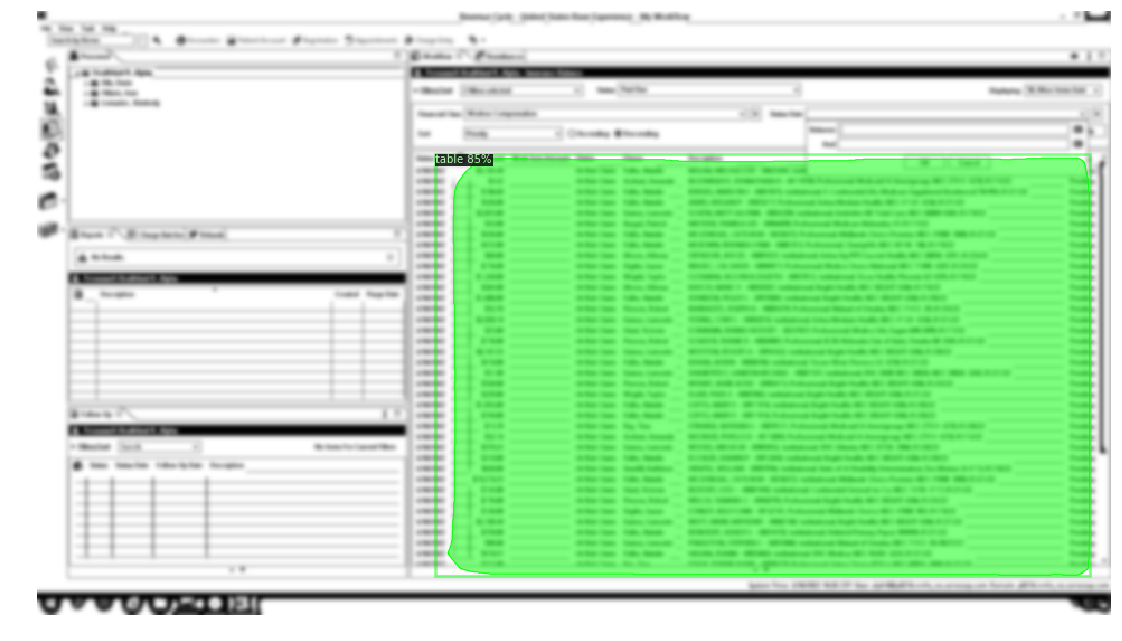}
      }
      \vspace{-1ex} 
      \setcounter{subfigure}{1}
      \subfigure[]{
        \includegraphics[width=0.4\textwidth]{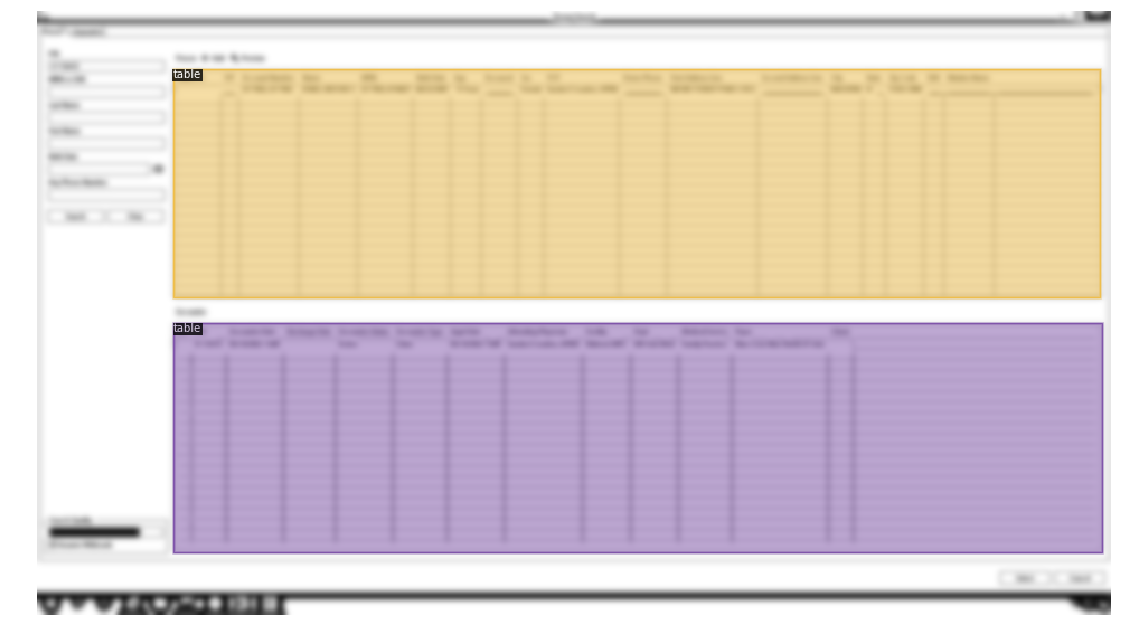}
        \includegraphics[width=0.4\textwidth]{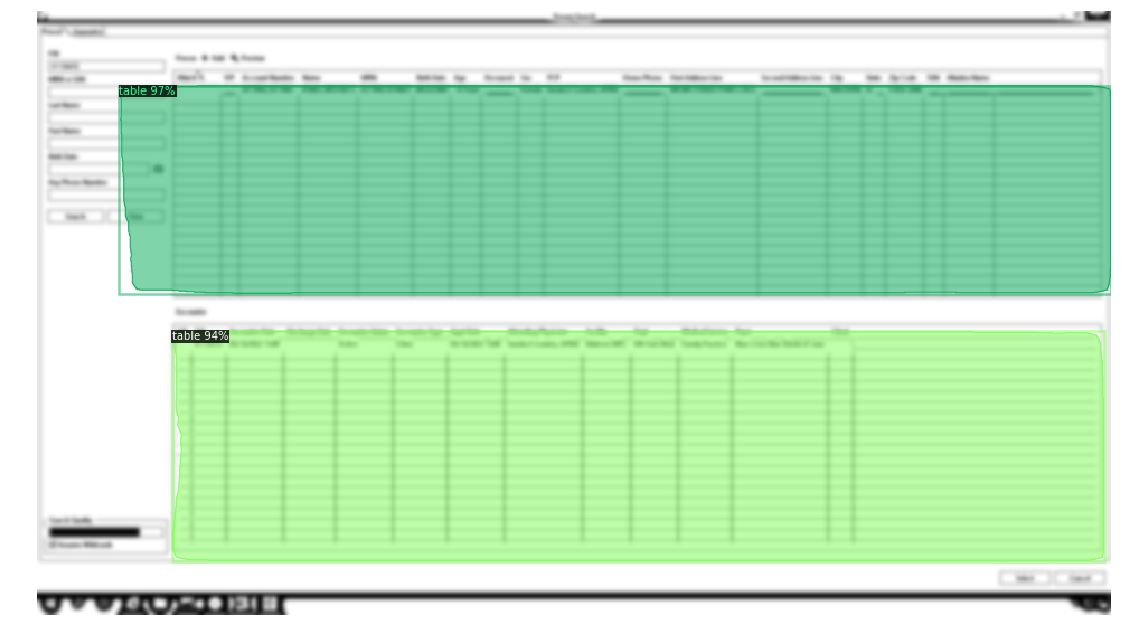}
      }
      \vspace{-1ex} 
      \setcounter{subfigure}{2}
      \subfigure[]{
        \includegraphics[width=0.4\textwidth]{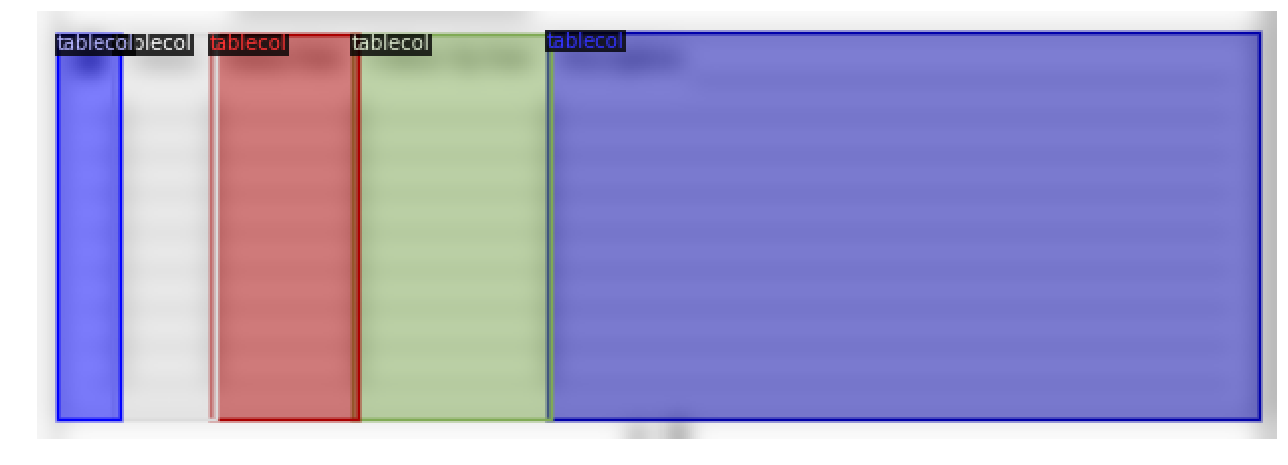}
        \includegraphics[width=0.4\textwidth]{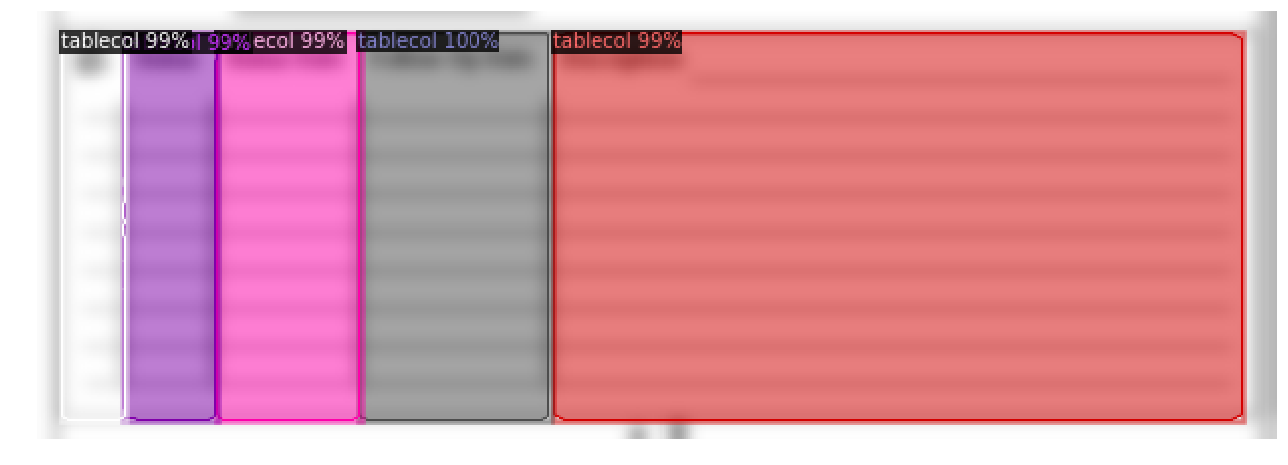}
      }
      \vspace{-1ex} 
      \setcounter{subfigure}{3}
      \subfigure[]{
        \includegraphics[width=0.4\textwidth]{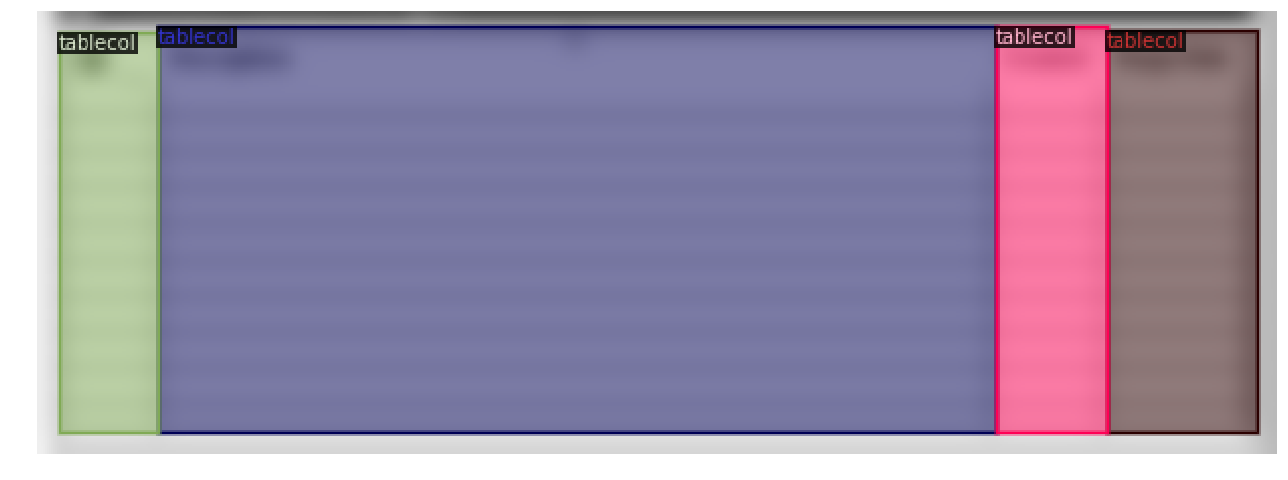}
        \includegraphics[width=0.4\textwidth]{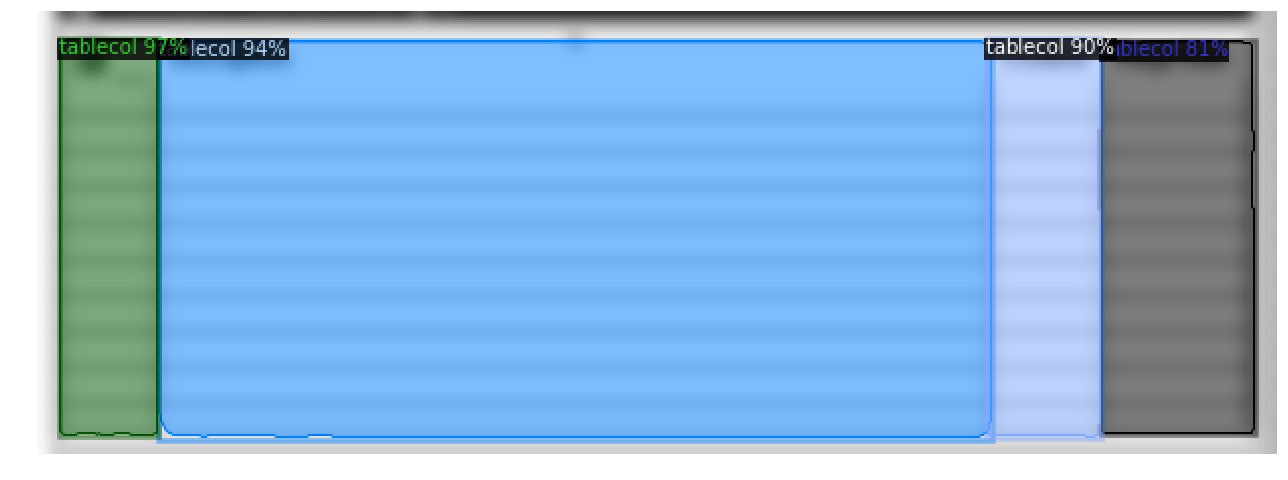}
      }
      \vspace{-1ex} 
      \setcounter{subfigure}{4}
      \subfigure[]{
        \includegraphics[width=0.4\textwidth]{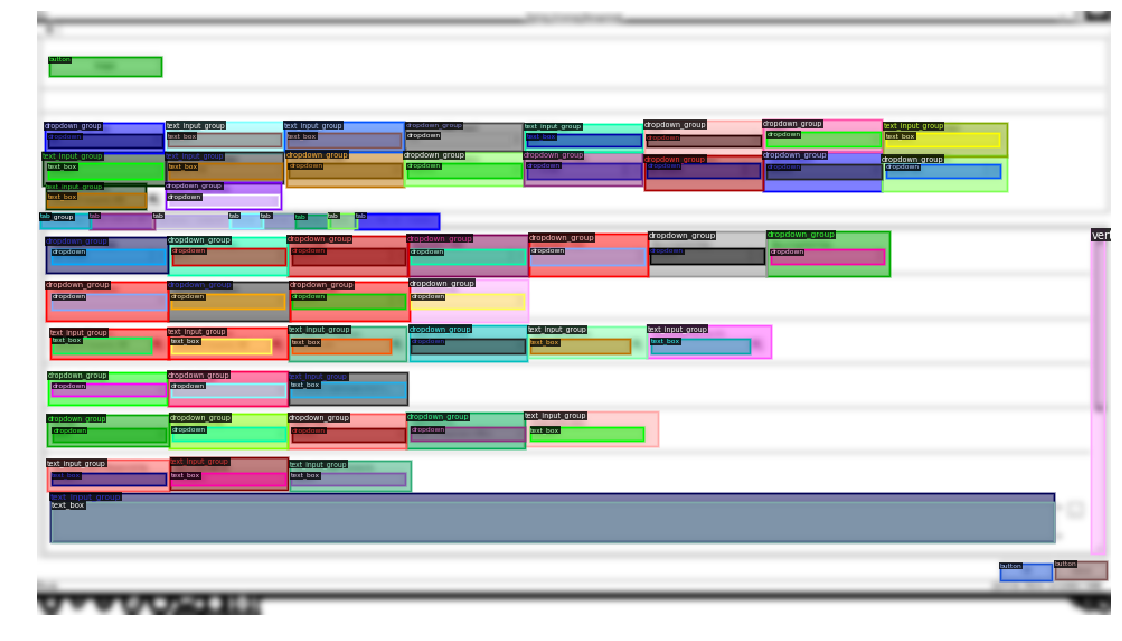}
        \includegraphics[width=0.4\textwidth]{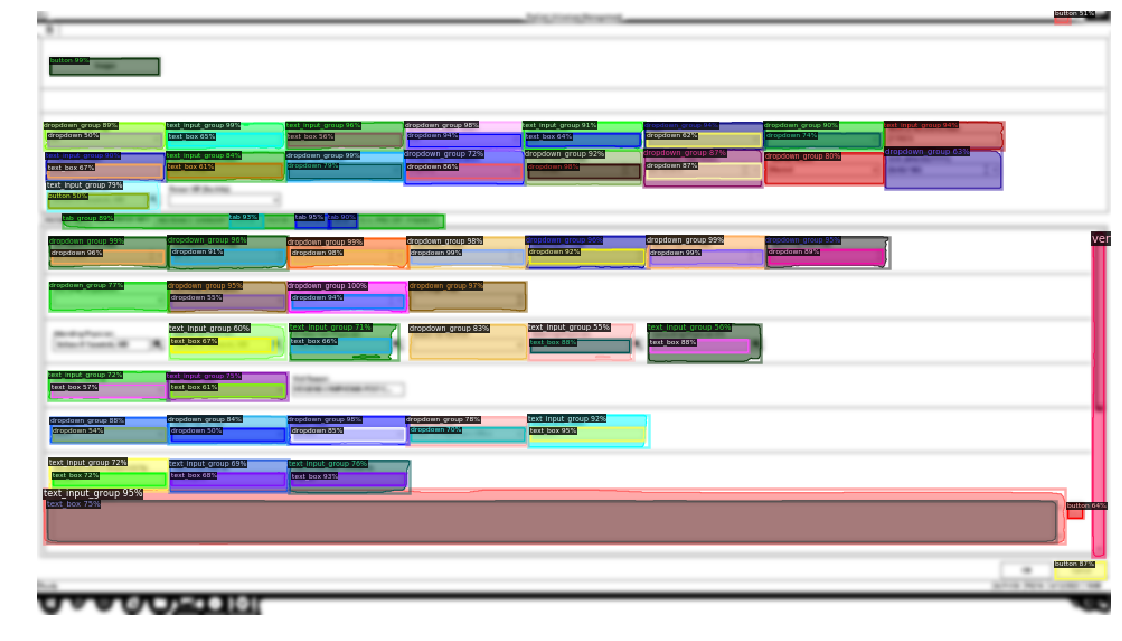}
      }
      \vspace{-1ex} 
      \setcounter{subfigure}{5}
      \subfigure[]{
        \includegraphics[width=0.4\textwidth]{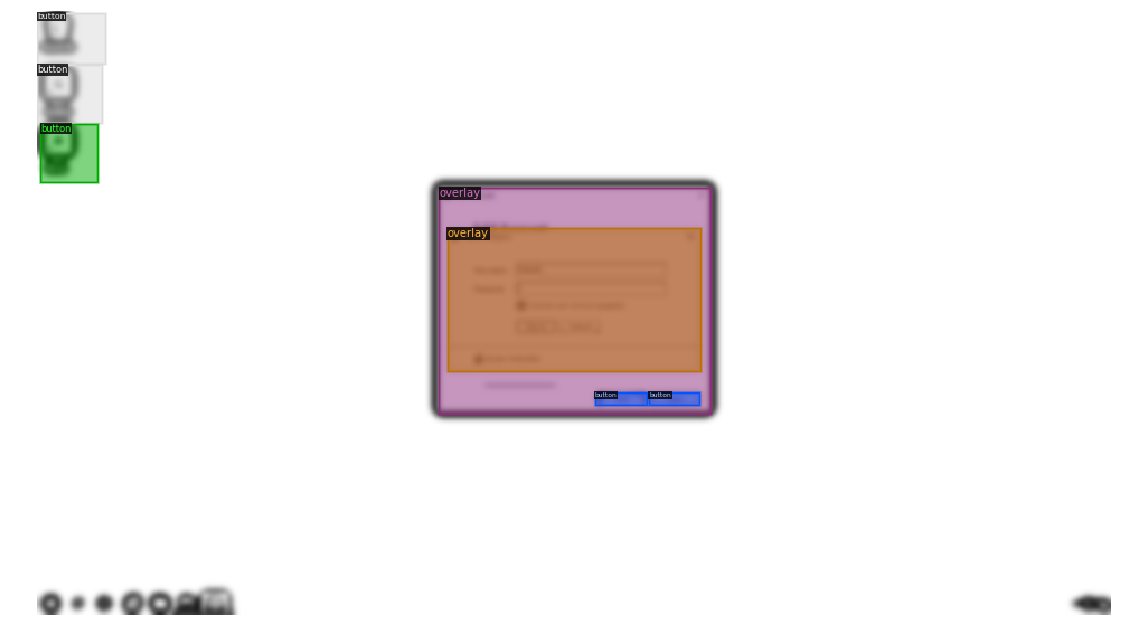}
        \includegraphics[width=0.4\textwidth]{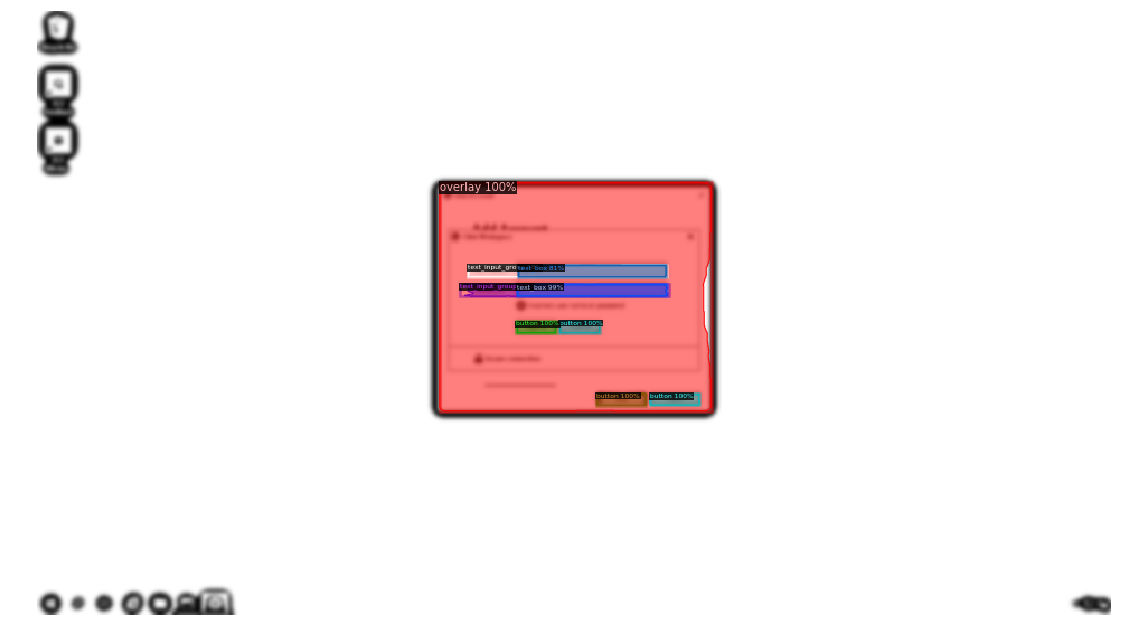}
      }
     \vspace{-1ex} 
     \caption{\textbf{\dataset~test set examples} (a-b) Table test set, (c-d) Column test set, and (e-f) GUI test set. The first column shows examples with human-annotated boundaries, while the second column presents the corresponding detection predictions by~\ours~. All images are purposefully scrubbed so that no PHI information about the patient is disclosed.}
     \label{fig:gui-example}
    \end{figure}

\textbf{\dataset~Dataset}
Beyond the TableBank dataset, we systematically curated an internal dataset of screen images from real-world EHR systems, which consists of screenshots collected by bots as they navigate the EHR systems of ten US health systems from various EHR providers.\footnote{The collection and use of this \dataset~dataset required HIPPA compliance training.} These screens were then carefully selected to include essential template screens and labeled by an internal team of labelers, including tables, table columns, and other GUI elements\footnote{Appendix~\ref{appendix:metadata} contains brief descriptions of all twelve classes of GUI elements labeled.}. Details of the dataset's composition can be found in Table~\ref{tab:datasets}. For the Column dataset, we further cropped these images to just tables, removing examples that were occluded from overlays. Visual examples of these datasets with human-annotated bounding boxes can be found in the first column of Figure~\ref{fig:gui-example}.  
    
In addition to the aforementioned supervised dataset with labeled objects, we also collected a \textit{unlabeled} Screenshot dataset from the worklogger system of a single US health system, which records videos of hospital staff interacting with the EHR system. We randomly sampled screen recordings from the database, then the randomly sampled frames from the randomly sampled videos. The frames were then split into two because most staff members had a dual monitor set up, and these frames were labeled by the same internal team of labelers, and frames without tables or columns or corrupted due to the splitting process were removed. This dataset contains 28,121 PNG images with a resolution of 1920x1080, which represent 10.8\% of the volume of the TableBank Train set. 
    
\subsection{Settings}
\textbf{Pretraining} For TableBank experiments, we pretrain~\ours~on the combined train set of 260,582 document images from the Word and Latex sets. For~\dataset~experiments, we have an unsupervised screenshot set of 28,121 images to be used for pretraining. For both experiments, we used adaptive binarization, random resize cropping, and random horizontal flipping as a weak augmentation to train the MAE branch. For a strong augmentation to train the eBT arm, we further add smudge transformation and dilation, which are introduced in~\citep{Prasad20}. The input images are resized to 224 by 224 in size, and the pretraining phase takes 500 epochs to complete with a batch size of 80. 

    \begin{table}[!ht]
        \centering
        \caption{\textbf{Results of table detection on the TableBank test set} (in AP and AP\textsubscript{75}). MAE denotes the representative SSL pretraining baseline, while ResNet~\citep{he2016deep} stands for the purely supervised baseline using the ResNet-152 backbone, with Cascade Mask R-CNN. The bold value represents the best (highest) value for each column metric. All baselines are outperformed by the proposed~\ours.}
        % \scriptsize
        \footnotesize
        \begin{tabular}{lrrrrr}\toprule
        &\multicolumn{2}{c}{\textbf{Word}} &\multicolumn{2}{c}{\textbf{Latex}} \\\cmidrule{2-5}
        \textbf{} &\textbf{AP} &\textbf{AP\textsubscript{75}} &\textbf{AP} &\textbf{AP\textsubscript{75}} \\\midrule
        \textbf{ResNet} (supervised baseline) &95.42 &95.78 &97.32 &98.62 \\
        \textbf{MAE} (self-supervised baseline) &95.94 &96.16 &97.63 &98.70 \\
        \textbf{\ours} (our method) &\textbf{96.03} &\textbf{96.22} &\textbf{97.68} &\textbf{98.75} \\
        \bottomrule
        \end{tabular}
        \label{tab:tb}
        \vspace{-1ex} 
    \end{table}
    
    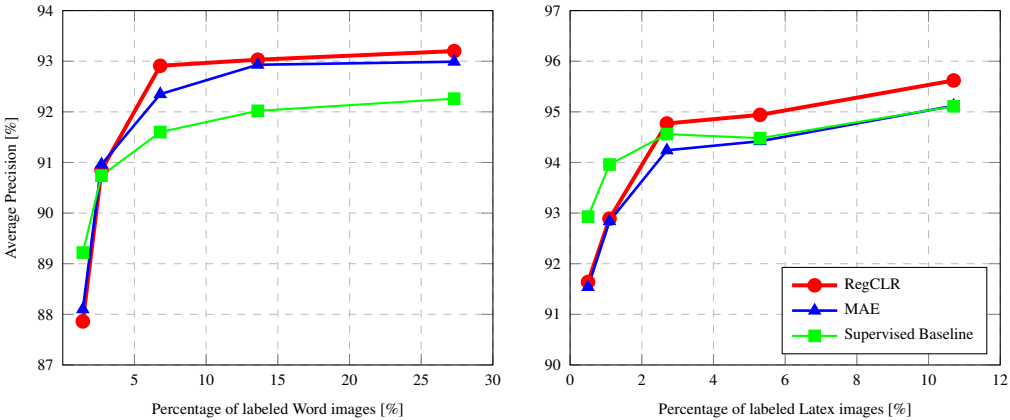
\begin{figure*}[!ht]
      \centering 
      \subfigure{
        \input{figures/plot_word.tikz}
      }
      \subfigure{
        \input{figures/plot_latex.tikz}
      }
      \caption{\textbf{Benefits of~\ours~in low labeled data regimes} (Left) Word set and (Right) Latex set. Each datapoint in the graph represents the number of labeled subsets of 1k, 2k, 5k, 10k, and 20k. \ours~begins performing best after a 2-3\% subset of labeled train data is provided for fine-tuning.}
      \label{fig:tb-ablation}  
      \vspace{-1ex} 
    \end{figure*}    

Our encoder and decoder model uses the ViT base backbone of a 12-layer Transformer with 12 self-attention heads. Our implementation of pretraining is built on the repo provided in MAE~\citep{He22}, and the projector is implemented as a 2 layer multilayer perceptron with 1024 dimensional output. Overall, pretraining on TableBank uses 4 Nvidia A100 GPUs for 20 hours, and pretraining on~\dataset~uses 4 Nvidia V100 GPUs for 40 hours. 

\textbf{Fine-tuning} For both experiments, we use MIMDet method to adapt pretrained ViT weights to detection. For all experiments except table detection in~\dataset~dataset, the input is resized so that the shortest side is between 480 and 800 pixels and the longest side is no more than 1333 pixels. For the table detection in~\dataset~dataset, the input is resized so that it has the shortest side between 307 and 512 pixels and the longest side no more than 853 pixels. We use adaptive binarization as preprocessing for input images, and our implementation for fine-tuning is based on the repo provided in MIMDet~\citep{fang2022unleashing}, with a 50\% sampling ratio for both training and inference. 
    
\textbf{Baselines} We carefully chose an MAE as a representative self-supervised pretraining baseline and ResNet with the ResNet-152 backbone as a fully supervised baseline. For a fair comparison, both baselines, like~\ours, use the Cascade Mask R-CNN architecture. All methods for TableBank datasets are trained for 125,000 iterations with a batch size of 12. All methods for table detection in the~\dataset~datasets are trained for 10,000 iterations with a batch size of 16 in the experiments. For other tasks in the~\dataset~datasets, we train for 20,000 iterations with a batch size of 4.

    \begin{table}[!ht]\centering
        \caption{\textbf{Results of GUI elements detection on the~\dataset~Table and Column test sets} (in AP and AP\textsubscript{75}). \ours~performs best when pretraining with the \dataset~Screenshot dataset, increasing AP scores relatively by 4.8\% for Table and 11.8\% for Column over the supervised baseline, as seen by comparing the first and second rows. Interestingly, despite pretraining with approximately 10\% volumes of TableBank, RegCLR fast approaches the best cross-domain transfer results from TableBank to EHRBank in the last row.}\label{tab:tabletablecol}
        \footnotesize
        % \scriptsize
        \begin{tabular}{lrrrrrrrr}\toprule
        & &\multicolumn{2}{c}{\textbf{Table}} &\multicolumn{2}{c}{\textbf{Column}} \\\cmidrule{3-6}
        Pretrain on &Method &\textbf{AP}  &\textbf{AP\textsubscript{75}} &\textbf{AP}  &\textbf{AP\textsubscript{75}} \\\midrule
        \multirow{1}{*}{N/A} 
        &\textbf{ResNet} &40.53 &44.46 &61.43 &67.07 \\\midrule
        \multirow{2}{*}{\dataset~Screenshot} 
        &\textbf{\ours} &\textbf{42.46} &\textbf{45.32} &\textbf{68.68} &\textbf{75.17} \\
        &\textbf{MAE} &36.78 &39.05 &64.67 &71.09 \\\midrule
        \multirow{2}{*}{TableBank (cross-domain)} 
        &\textbf{\ours} &40.96 &43.47 &67.77  &75.06 \\
        &\textbf{MAE} &\textbf{43.99} &\textbf{48.77} &\textbf{69.83} &\textbf{77.29} \\
        \bottomrule
        \end{tabular}
        \vspace{-1ex} 
    \end{table}
    
    \begin{table}[!ht]
        \caption{\textbf{Results of GUI elements detection on the~\dataset~GUI test set} (Left) GUI elements detection AP and AP\textsubscript{75} in the GUI test set, and (Right) the detection AP of each GUI element class in the test set. As shown in the left table, \ours~outperforms the baselines in both AP and AP\textsubscript{75} metrics on average. Specifically, \ours~outperforms the baselines in eight of twelve categories, while MAE performs even worse than ResNet in six.}\label{tab:gui}
        \begin{subtable}
        \centering
            \footnotesize
            % \scriptsize
            \begin{tabular}{lrrrr}\toprule
            &\textbf{ResNet} &\textbf{MAE} &\textbf{RegCLR} \\\midrule
            \textbf{AP} &43.37 &45.69 &\textbf{48.17} \\
            \textbf{AP\textsubscript{75}} &48.25 &50.80 &\textbf{54.10} \\
            \bottomrule
            \end{tabular}
        \end{subtable}
        % \hspace{\fill}
        \hfill
        \begin{subtable}
        \centering
            \footnotesize
            % \scriptsize
            \begin{tabular}{lrrrr}\toprule
                &\textbf{ResNet} &\textbf{MAE} &\textbf{\ours} \\\midrule
                \textbf{Button} &\textbf{38.42} &33.02 &33.13 \\
                \textbf{Dropdown} &\textbf{48.02} &44.33 &43.14 \\
                \textbf{Dropdown\_group} &43.25 &41.71 &\textbf{43.33} \\
                \textbf{Horizontal\_scrollbar} &25.12 &36.88 &\textbf{38.78} \\
                \textbf{Overlay} &52.92 &\textbf{65.58} &62.12 \\
                \textbf{Tab} &\textbf{37.62} &34.13 &36.46 \\
                \textbf{Tab\_group} &29.94 &30.35 &\textbf{32.77} \\
                \textbf{Table} &62.72 &66.10 &\textbf{72.64} \\
                \textbf{Table\_column} &46.31 &50.44 &\textbf{54.74} \\
                \textbf{Text\_box} &55.03 &54.56 &\textbf{58.06} \\
                \textbf{Text\_input\_group} &42.82 &41.38 &\textbf{44.87} \\
                \textbf{Vertical\_scrollbar} &37.28 &47.32 &\textbf{52.00} \\
                \bottomrule
            \end{tabular}
        \end{subtable}
        \vspace{-1ex} 
    \end{table}

\subsection{Results}
\textbf{TableBank Dataset}
% Table detection prediction results (in AP and AP\textsubscript{75}) on the TableBank test set. MAE denotes the representative SSL baseline, while ResNet stands for the suvervised only baseline using the ResNet-152 backbone, with Cascade Mask R-CNN. The bold value represents the best (highest) value for each column metric. All baselines are outperformed by the proposed~\ours.
The prediction results of the TableBank table detection are shown in Table~\ref{tab:tb} in AP (mAP @ IOU [0.50:0.95]) and AP\textsubscript{75} (mAP @ IOU 0.75) with the results of two baselines. We can see that our method~\ours~outperforms the other self-supervised and fully supervised baselines. The visualization of the table detection predictions by~\ours~can be found in the second row of Figure~\ref{fig:tb-example}. We conduct further evaluations to better understand our method,~\ours, especially how it performs when the number of labeled data decreases, particularly in low labeled data regimes. 

To do so, we evaluate the test set when fine-tuning on the subsets of 1k, 2k, 5k, 10k, and 20k. Note that all methods are trained for 10,000 iterations with a batch size of 12 for this experiment. As shown in Figure~\ref{fig:tb-ablation},~\ours~outperforms the baselines soon after fine-tuning on a subset of 2-3\% labeled train set, AP scores improve quickly even with considerably fewer labeled images (i.e., less than 10\%) provided than the unlabeled set, and the gain decreases as more labeled data are added. 

\textbf{\dataset~Dataset}
We then evaluate \ours~on internal~\dataset~dataset which has higher values in our production scenario. As shown in Table~\ref{tab:tabletablecol}, when pretrained on \textit{unlabeled}~\dataset, our method outperforms the baselines in both Table and Column detection, increasing relative AP scores by 4.8\% and 11.8\% respectively over the supervised baseline, as seen by comparing the first and second rows. Furthermore, even though pretrained with only around 10\% of the TableBbank volume, \ours~quickly approaches the best cross-domain transfer performance from TableBank to EHRBank, as shown in the last row. 

Additionally, it is worth noting that MAE performs worse than even ResNet on Table when pretrained on~\dataset~(by comparing the first and third rows). It does, however, transfer better than~\ours~in scenarios involving cross-domain transfer from public TableBank to private~\dataset~ (by comparing the last two rows). In the future, we intend to investigate how quickly detection performance improves as the unlabeled data volume scales, as well as how effectively pretrained weights transfer across domains in the context of tabular rich images, so that they can be applicable to other document format datasets (e.g., Word to GUI and vice versa). 

Table~\ref{tab:gui} presents the results in GUI elements detection, in which our method pretrained on~\dataset~Screenshot again produces the highest overall detection scores compared to the baselines. This demonstrates that~\ours~can provide performance improvement even in more complicated scenarios with a larger number of classes and possible occlusions between different GUI elements. The performance variation across different GUI categories is also presented in the right table of Table~\ref{tab:gui}. Visualization of the predictions in the~\dataset~can be found in Figure~\ref{fig:gui-example}. 

    % \begin{figure}[ht]\centering
    %     \sidesubfloat[]{\includegraphics[width=2in]{figures/gt-table-1.png}\label{fig:gui-a-gt}}
    %     \sidesubfloat{\includegraphics[width=2in]{figures/det-table-1.png}\label{fig:gui-a-det}}
    %     \setcounter{subfigure}{1}
    %     \sidesubfloat[]{\includegraphics[width=2in]{figures/gt-table-2.png}\label{fig:gui-b-gt}}
    %     \sidesubfloat{\includegraphics[width=2in]{figures/det-table-2.png}\label{fig:gui-b-det}}
    %     \setcounter{subfigure}{2}
    %     \sidesubfloat[]{\includegraphics[width=2in]{figures/gt-tablecol-1.png}\label{fig:gui-c-gt}}
    %     \sidesubfloat{\includegraphics[width=2in]{figures/det-tablecol-1.png}\label{fig:gui-c-det}}
    %     \setcounter{subfigure}{3}
    %     \sidesubfloat[]{\includegraphics[width=2in]{figures/gt-tablecol-2.png}\label{fig:gui-d-gt}}
    %     \sidesubfloat{\includegraphics[width=2in]{figures/det-tablecol-2.png}\label{fig:gui-d-det}}
    %     \setcounter{subfigure}{4}
    %     \sidesubfloat[]{\includegraphics[width=2in]{figures/gt-gui-1.png}\label{fig:gui-e-gt}}
    %     \sidesubfloat{\includegraphics[width=2in]{figures/det-gui-1.png}\label{fig:gui-e-det}}
    %     \setcounter{subfigure}{5}
    %     \sidesubfloat[]{\includegraphics[width=2in]{figures/gt-gui-2.png}\label{fig:gui-f-gt}}
    %     \sidesubfloat{\includegraphics[width=2in]{figures/det-gui-2.png}\label{fig:gui-f-det}}
    %     \caption{Internal dataset and Detection prediction for internal dataset examples. (a-d) Internal table dataset examples; (e-h) Internal table column dataset examples, (i-l): Internal GUI elements dataset examples.}
    %     \label{fig:gui-example}
    % \end{figure}

%% file: figures/plot_word.tikz
% \documentclass{standalone}
% \usepackage{tikz}
% \usepackage{pgfplots}
% \pgfplotsset{width=10.0cm,compat=newest}

% \begin{document}
\begin{tikzpicture}

\begin{axis}[
    xlabel={Percentage of labeled Word images [\%]}, 
    ylabel={Average Precision [\%]},
    xmin=0.0, xmax=30.0, xtick={5,10,...,25,30},
    ymin=87, ymax=94, ytick={87,88,...,93,94},    
    xmajorgrids=true, 
    ymajorgrids=true, 
    grid style=dashed,
    label style={font=\tiny}, 
    tick label style={font=\tiny},  
    legend pos=south east,
    legend style={font=\tiny},
    legend cell align={left}
]

\addplot[color=red,mark=*,line width=1.5pt]
    coordinates {(1.4,87.86)
                (2.7,90.85)
                (6.8,92.91)
                (13.6,93.03)
                (27.3,93.20)
};

\addplot[color=blue,mark=triangle*,line width=1.0pt]
    coordinates {(1.4,88.10)
                (2.7,90.96)
                (6.8,92.35)
                (13.6,92.93)
                (27.3,92.99)
};

\addplot[color=green,mark=square*,line width=0.8pt]
    coordinates {(1.4,89.22)
                (2.7,90.74)
                (6.8,91.60)
                (13.6,92.02)
                (27.3,92.26)
};

% \legend{RegCLR, MAE, Supervised Baseline}
\end{axis}

\end{tikzpicture}
% \end{document}

%% file: figures/plot_latex.tikz
% \documentclass{standalone}
% \usepackage{tikz}
% \usepackage{pgfplots}
% \pgfplotsset{width=10.0cm,compat=newest}

% \begin{document}
\begin{tikzpicture}

\begin{axis}[
    xlabel={Percentage of labeled Latex images [\%]}, 
    xmin=0.0, xmax=12.0, xtick={0,2,...,10,12},
    ymin=90, ymax=97, ytick={90,91,...,96,97},    
    xmajorgrids=true, 
    ymajorgrids=true, 
    grid style=dashed,
    label style={font=\tiny}, 
    tick label style={font=\tiny},  
    legend pos=south east,
    legend style={font=\tiny},
    legend cell align={left}
]

\addplot[color=red,mark=*,line width=1.5pt]
    coordinates {(0.5,91.64)
                (1.1,92.89)
                (2.7,94.77)
                (5.3,94.94)
                (10.7,95.62)
};

\addplot[color=blue,mark=triangle*,line width=1.0pt]
    coordinates {(0.5,91.54)
                (1.1,92.84)
                (2.7,94.24)
                (5.3,94.42)
                (10.7,95.12)
};

\addplot[color=green,mark=square*,line width=0.8pt]
    coordinates {(0.5,92.93)
                (1.1,93.96)
                (2.7,94.56)
                (5.3,94.48)
                (10.7,95.11)
};

\legend{RegCLR, MAE, Supervised Baseline}
\end{axis}

\end{tikzpicture}
% \end{document}

%% file: sections/20_related_work.tex
\section{Related Work}
\label{sec:relatedwork}

\textbf{SSL Methods for Natural Images} Learning good visual representation in a self-supervised manner has been a rising paradigm in computer vision over the past few years. More recently, a few families of approaches have emerged. i) Contrastive learning is trained to extract meaningful features by mapping different images to distant points and the same image under various image augmentations to close points in the feature space. As an example, SimCLR~\citep{Chen20} defines the feature distance by cosine similarity and utilizes InfoNCE proposed by~\citep{oord2018representation} as a loss function. ii) Self-distillation serves as another powerful approach that achieves effective representation learning while alleviating the need for large contrastive batch sizes that are often required for contrastive methods. BYOL~\citep{grill2020bootstrap} simply minimizes the similarity loss between the encoder and its momentum target version to achieve this. iii) Feature regularization methods, alternatively, optimize over the auxiliary regularizing loss to directly learn features with desired properties. In this family, Barlow Twins~\citep{zbontar2021barlow} minimizes off-diagonal terms in the covariance matrix between two augmentations to obtain low redundancy representations.

\textbf{Transformer Architecture for SSL Methods} In addition to the advancement of the SSL algorithm, the Transformer~\citep{vaswani2017attention} architecture also presents itself as a powerful general architecture in computer vision. It brings advantages over CNN, which has dominated the field as a de facto network structure since the emergence of deep learning. Using the ViT architecture, MAE~\citep{He22}, in particular, generalizes masked language modeling (MLM) popular in natural language processing to MIM in computer vision. MAE feeds unmasked patches to the transformer and predicts masked patches as a pretext task. Despite its simplicity, MAE shows strong performance on various computer vision tasks, and specifically, MAE has been found to produce features that better support the downstream object detection task~\citep{li2021benchmarking} that was the inspiration for our study.

% It has been shown that the Transformer architecture is well-suited for SSL paradigm. Using momentum encoder and multi-crop augmentation, DINO~\citep{caron2021emerging} shows that the self-supervisedly pretrained Vision Transformers (ViT)~\citep{dosovitskiy2020image} can generate features that embeds explicit information to the semantic segmentation which has not been found in either self-supervised CNN or supervised ViT. 
% Using ViT architecture, MAE~\citep{He22}, on the other hand, generalizes the masked language modeling (MLM) popular in natural language processing to MIM in computer vision. MAE feeds unmasked patches to the transformer and predicts masked patches as a pretext task. Despite its simplicity, MAE shows strong performance in various computer vision tasks. In particular, MAE has been found to produce features that better support the downstream object detection task~\citep{li2021benchmarking} which was the inspiration for our study. Followup work~\citep{fang2022unleashing} finds that the MAE decoder can serve as a complimentary feature extractor has encoder to further enhance performance of pretrained MAE for object detection.

\textbf{SSL Methods for Document Images} Beyond natural images, self-supervised pre-training with transformers is actively used in vision-based document AI tasks. In particular, DiT~\citep{li2022dit}, which uses BEiT~\citep{bao2021beit}, demonstrates MIM's potential to extract high-quality visual representations from document images. LayoutLMv3~\citep{huang2022layoutlmv3} further extends pretraining to a multimodal setup with unified text and image masking, including MLM for texts, MIM for images, and a novel word-patch alignment that predicts whether the corresponding image patch for any given word is masked. In the direction of multimodal pretraining for document images, VLCDoC~\citep{bakkali2022vlcdoc} proposes using contrastive loss to align text and vision cues, producing strong performance when both text and image are available. In the context of the images of medical-health documents that we are interested in, DEXTER~\citep{pr2022dexter} fine-tunes the TableBank from pretrained MS COCO weights, while no SSL method is employed.

%% file: sections/60_conclusion.tex
\section{Conclusion}
\label{sec:conclusion}
In this paper, we present~\ours, a novel self-supervised framework for learning tabular representation and downstream detections, which had received little attention than natural image domain. Our brand-new framework combines contrastive and regularized self-supervised methods and has been pretrained on both public and private domain tabular rich images. We demonstrate that~\ours~outperforms previous self-supervised pretraining and fully supervised baselines by a large margin in various real-world contexts, with high sample efficiency for fine-tuning. It is left to future study how quickly the quality of the representation learned by our method improves as the volume of \textit{unlabeled} images scales, which is a common and practical scenario in production. We believe that this study is an important step towards semantic comprehension of real-world document images, and it will be interesting to see how this vision-based framework can be expanded to include textual content without manual data annotation.

% In this paper, we present~\ours, a novel self-supervised framework for learning tabular representation and downstream detections, which had received little attention than natural images. Our brand-new framework combines contrastive and regularized self-supervised methods and has been pretrained on both public and private domain tabular rich image datasets. Experimental results demonstrate that~\ours~framework outperforms previous self-supervised pretraining and fully supervised baselines by a large margin across various real-world contexts, with high sample efficiency for fine-tuning. We have not discussed how quickly the quality of the tabular representation learned by our method improves as the volume of \textit{unlabeled} images scales, which is highly common and practical in the production scenario and has been left for future study. However, we believe that this~\ours~is a significant step towards the goal of complete understanding of real-world document images (e.g., semantic understanding of GUI screens and components), and it will be interesting to see how this vision-based framework can be extended to integrate textual information without the need for manual data annotation efforts.

%% file: sections/70_ack.tex
The authors express their gratitude to everyone who contributed to the EHRBank dataset, especially Aaron Loh, Fernando Silveira, Samuel Masling, Ihsaan Patel, and the internal labeling teams. Angela Kilby and Jiaming Zeng's suggestions to improve the manuscripts are also sincerely acknowledged.

%% file: sections/100_appendix.tex
% \pagebreak
\section{Appendix}

\textbf{\dataset~GUI Dataset:} We collect and label GUI screenshots with twelve classes: button, dropdown, drowdown group, overlay, tab, tab group, table, table column, text box, text input group, horizontal scrollbar and vertical scrollbar. 
\begin{itemize}
    \item Button: Small, usually rectangle shape, clickable elements in the screen with or without text.
    \item Overlay: A separate piece of user interface that appears to be the front layer (e.g., a popup window).
    \item Dropdown and dropdown group: A clickable rectangle shape UI element or a group of such UI elements that can be clicked to expand to multiple choice selection interface. 
    \item Tab and tab group: A clickable rectangle shape UI element or a group of such UI elements that can switch between different pages.
    \item Table and table column: Tabular element and column of such tabular element.
    \item Text box and text input group: Box that can type in text and group of such elements.
    \item Horizontal scrollbar and vertical: scrollbar: Horizontal and vertical draggable UI element that can move the displaying window accordingly.
    
\end{itemize}
\label{appendix:metadata}

%% file: neurips_2022.bbl
\begin{thebibliography}{28}
\providecommand{\natexlab}[1]{#1}
\providecommand{\url}[1]{\texttt{#1}}
\expandafter\ifx\csname urlstyle\endcsname\relax
  \providecommand{\doi}[1]{doi: #1}\else
  \providecommand{\doi}{doi: \begingroup \urlstyle{rm}\Url}\fi

\bibitem[Agarwal et~al.(2021)Agarwal, Mondal, and Jawahar]{Agarwal21}
M.~Agarwal, A.~Mondal, and C.~V. Jawahar.
\newblock Cdec-net: Composite deformable cascade network for table detection in
  document images.
\newblock In \emph{2020 25th International Conference on Pattern Recognition
  (ICPR)}, 2021.

\bibitem[Bakkali et~al.(2022)Bakkali, Ming, Coustaty, Rusi{\~n}ol, and
  Terrades]{bakkali2022vlcdoc}
S.~Bakkali, Z.~Ming, M.~Coustaty, M.~Rusi{\~n}ol, and O.~R. Terrades.
\newblock Vlcdoc: Vision-language contrastive pre-training model for
  cross-modal document classification.
\newblock \emph{arXiv preprint arXiv:2205.12029}, 2022.

\bibitem[Bao et~al.(2021)Bao, Dong, and Wei]{bao2021beit}
H.~Bao, L.~Dong, and F.~Wei.
\newblock Beit: Bert pre-training of image transformers.
\newblock \emph{arXiv preprint arXiv:2106.08254}, 2021.

\bibitem[Burdick et~al.(2020)Burdick, Evfimievski, Katsis, Danilevsky, and
  Wang]{Burdick20}
D.~Burdick, A.~V. Evfimievski, Y.~Katsis, M.~Danilevsky, and N.~Wang.
\newblock Ibm's tutorial on table extraction and understanding for scientific
  and enterprise applications, 2020.
\newblock URL
  \url{https://researcher.watson.ibm.com/researcher/view_group.php?id=10211}.

\bibitem[Cai and Vasconcelos(2019)]{Cai_2019}
Z.~Cai and N.~Vasconcelos.
\newblock Cascade r-cnn: High quality object detection and instance
  segmentation.
\newblock \emph{IEEE Transactions on Pattern Analysis and Machine
  Intelligence}, page 1–1, 2019.
\newblock ISSN 1939-3539.
\newblock \doi{10.1109/tpami.2019.2956516}.
\newblock URL \url{http://dx.doi.org/10.1109/tpami.2019.2956516}.

\bibitem[Chen et~al.(2020)Chen, Kornblith, Norouzi, and Hinton]{Chen20}
T.~Chen, S.~Kornblith, M.~Norouzi, and G.~Hinton.
\newblock A simple framework for contrastive learning of visual
  representations.
\newblock In \emph{International conference on machine learning}, pages
  1597--1607. PMLR, 2020.

\bibitem[Dosovitskiy et~al.(2021)Dosovitskiy, Beyer, Kolesnikov, Weissenborn,
  Zhai, Unterthiner, Dehghani, Minderer, Heigold, Gelly, Uszkoreit, and
  Houlsby]{Dosovitskiy21}
A.~Dosovitskiy, L.~Beyer, A.~Kolesnikov, D.~Weissenborn, X.~Zhai,
  T.~Unterthiner, M.~Dehghani, M.~Minderer, G.~Heigold, S.~Gelly, J.~Uszkoreit,
  and N.~Houlsby.
\newblock An image is worth 16x16 words: Transformers for image recognition at
  scale.
\newblock In \emph{9th International Conference on Learning Representations
  (ICLR 2021)}, 2021.

\bibitem[Fang et~al.(2022)Fang, Yang, Wang, Ge, Shan, and
  Wang]{fang2022unleashing}
Y.~Fang, S.~Yang, S.~Wang, Y.~Ge, Y.~Shan, and X.~Wang.
\newblock Unleashing vanilla vision transformer with masked image modeling for
  object detection.
\newblock \emph{arXiv preprint arXiv:2204.02964}, 2022.

\bibitem[Gidaris et~al.(2018)Gidaris, Singh, and
  Komodakis]{gidaris2018unsupervised}
S.~Gidaris, P.~Singh, and N.~Komodakis.
\newblock Unsupervised representation learning by predicting image rotations.
\newblock \emph{arXiv preprint arXiv:1803.07728}, 2018.

\bibitem[Grill et~al.(2020)Grill, Strub, Altch{\'e}, Tallec, Richemond,
  Buchatskaya, Doersch, Avila~Pires, Guo, Gheshlaghi~Azar,
  et~al.]{grill2020bootstrap}
J.-B. Grill, F.~Strub, F.~Altch{\'e}, C.~Tallec, P.~Richemond, E.~Buchatskaya,
  C.~Doersch, B.~Avila~Pires, Z.~Guo, M.~Gheshlaghi~Azar, et~al.
\newblock Bootstrap your own latent-a new approach to self-supervised learning.
\newblock \emph{Advances in neural information processing systems},
  33:\penalty0 21271--21284, 2020.

\bibitem[He et~al.(2016)He, Zhang, Ren, and Sun]{he2016deep}
K.~He, X.~Zhang, S.~Ren, and J.~Sun.
\newblock Deep residual learning for image recognition.
\newblock In \emph{Proceedings of the IEEE conference on computer vision and
  pattern recognition}, pages 770--778, 2016.

\bibitem[He et~al.(2022)He, Chen, Xie, Li, Dollár, and Girshick]{He22}
K.~He, X.~Chen, S.~Xie, Y.~Li, P.~Dollár, and R.~Girshick.
\newblock Masked autoencoders are scalable vision learners.
\newblock In \emph{Proceedings of Computer Vision and Pattern Recognition (CVPR
  2022)}. IEEE, 2022.

\bibitem[Huang et~al.(2022)Huang, Lv, Cui, Lu, and Wei]{huang2022layoutlmv3}
Y.~Huang, T.~Lv, L.~Cui, Y.~Lu, and F.~Wei.
\newblock Layoutlmv3: Pre-training for document ai with unified text and image
  masking.
\newblock \emph{arXiv preprint arXiv:2204.08387}, 2022.

\bibitem[LeCun(2022)]{LeCun22}
Y.~LeCun.
\newblock A path towards autonomous machine intelligence.
\newblock \emph{OpenReview preprint}, 2022.

\bibitem[Lewis et~al.(2006)Lewis, Agam, Argamon, Frieder, Grossman, and
  Heard]{lewis2006building}
D.~Lewis, G.~Agam, S.~Argamon, O.~Frieder, D.~Grossman, and J.~Heard.
\newblock Building a test collection for complex document information
  processing.
\newblock In \emph{Proceedings of the 29th annual international ACM SIGIR
  conference on Research and development in information retrieval}, pages
  665--666, 2006.

\bibitem[Li et~al.(2022)Li, Xu, Lv, Cui, Zhang, and Wei]{li2022dit}
J.~Li, Y.~Xu, T.~Lv, L.~Cui, C.~Zhang, and F.~Wei.
\newblock Dit: Self-supervised pre-training for document image transformer.
\newblock \emph{arXiv preprint arXiv:2203.02378}, 2022.

\bibitem[Li et~al.(2019)Li, Cui, Huang, Wei, Zhou, and Li]{li2019tablebank}
M.~Li, L.~Cui, S.~Huang, F.~Wei, M.~Zhou, and Z.~Li.
\newblock Tablebank: A benchmark dataset for table detection and recognition.
\newblock \emph{arXiv preprint arXiv:1903.01949}, 2019.

\bibitem[Li et~al.(2021)Li, Xie, Chen, Dollar, He, and
  Girshick]{li2021benchmarking}
Y.~Li, S.~Xie, X.~Chen, P.~Dollar, K.~He, and R.~Girshick.
\newblock Benchmarking detection transfer learning with vision transformers.
\newblock \emph{arXiv preprint arXiv:2111.11429}, 2021.

\bibitem[Noroozi and Favaro(2016)]{noroozi2016unsupervised}
M.~Noroozi and P.~Favaro.
\newblock Unsupervised learning of visual representations by solving jigsaw
  puzzles.
\newblock In \emph{European conference on computer vision}, pages 69--84.
  Springer, 2016.

\bibitem[Oord et~al.(2018)Oord, Li, and Vinyals]{oord2018representation}
A.~v.~d. Oord, Y.~Li, and O.~Vinyals.
\newblock Representation learning with contrastive predictive coding.
\newblock \emph{arXiv preprint arXiv:1807.03748}, 2018.

\bibitem[PR et~al.(2022)PR, Krishnamoorthy, Srivatsan, Goyal, and
  Santhiappan]{pr2022dexter}
N.~PR, H.~Krishnamoorthy, K.~Srivatsan, A.~Goyal, and S.~Santhiappan.
\newblock Dexter: An end-to-end system to extract table contents from
  electronic medical health documents.
\newblock \emph{arXiv preprint arXiv:2207.06823}, 2022.

\bibitem[Prasad et~al.(2020)Prasad, Gadpal, Kapadni, Visave, and
  Sultanpure]{Prasad20}
D.~Prasad, A.~Gadpal, K.~Kapadni, M.~Visave, and K.~Sultanpure.
\newblock Cascadetabnet: An approach for end to end table detection and
  structure recognition from image-based documents.
\newblock In \emph{CVPR 2020 Workshop on Text and Documents in the Deep
  Learning Era}, 2020.

\bibitem[Slonim et~al.(2006)Slonim, Friedman, and Tishby]{Slonim06}
N.~Slonim, N.~Friedman, and N.~Tishby.
\newblock {Multivariate Information Bottleneck}.
\newblock \emph{Neural Computation}, 18\penalty0 (8):\penalty0 1739--1789, 08
  2006.

\bibitem[Vashisth et~al.(2022)Vashisth, Mullen, Emmott, and Lozada]{Vashisth22}
S.~Vashisth, A.~Mullen, S.~Emmott, and A.~Lozada.
\newblock Gartner market guide for intelligent document processing solutions,
  2022.

\bibitem[Vaswani et~al.(2017)Vaswani, Shazeer, Parmar, Uszkoreit, Jones, Gomez,
  Kaiser, and Polosukhin]{vaswani2017attention}
A.~Vaswani, N.~Shazeer, N.~Parmar, J.~Uszkoreit, L.~Jones, A.~N. Gomez,
  {\L}.~Kaiser, and I.~Polosukhin.
\newblock Attention is all you need.
\newblock \emph{Advances in neural information processing systems}, 30, 2017.

\bibitem[Vincent et~al.(2008)Vincent, Larochelle, Bengio, and
  Manzagol]{Vincent08}
P.~Vincent, H.~Larochelle, Y.~Bengio, and P.-A. Manzagol.
\newblock Extracting and composing robust features with denoising autoencoders.
\newblock \emph{Technical Report}, 2008.

\bibitem[Zbontar et~al.(2021)Zbontar, Jing, Misra, LeCun, and
  Deny]{zbontar2021barlow}
J.~Zbontar, L.~Jing, I.~Misra, Y.~LeCun, and S.~Deny.
\newblock Barlow twins: Self-supervised learning via redundancy reduction.
\newblock In \emph{International Conference on Machine Learning}, pages
  12310--12320. PMLR, 2021.

\bibitem[Zhang et~al.(2016)Zhang, Isola, and Efros]{zhang2016colorful}
R.~Zhang, P.~Isola, and A.~A. Efros.
\newblock Colorful image colorization.
\newblock In \emph{European conference on computer vision}, pages 649--666.
  Springer, 2016.

\end{thebibliography}
